\documentclass[10pt,twocolumn,letterpaper]{article}

\usepackage{cvpr}
\usepackage{times}
\usepackage{epsfig}
\usepackage{graphicx}
\usepackage{amsmath}
\usepackage{amssymb}
\usepackage{multirow}
\usepackage{booktabs}
\usepackage{subcaption}
\usepackage{multirow}
\usepackage{balance}
\newcommand{\shape}{\mbox{\boldmath $\beta$}}
\newcommand{\pose}{\mbox{\boldmath $\theta$}}

\newcommand{\cam}{\mbox{\boldmath $\pi$}}

\newcommand\blfootnote[1]{%
  \begingroup
  \renewcommand\thefootnote{}\footnote{#1}%
  \addtocounter{footnote}{-1}%
  \endgroup
}

\usepackage[pagebackref=true,breaklinks=true,letterpaper=true,colorlinks,bookmarks=false]{hyperref}

\cvprfinalcopy 

\def\cvprPaperID{****} 

\ifcvprfinal\fi
\begin{document}

\title{Coherent Reconstruction of Multiple Humans from a Single Image}

\author{Wen Jiang$^2$\textsuperscript{*}, Nikos Kolotouros$^1$\textsuperscript{*}, Georgios Pavlakos$^1$, Xiaowei Zhou$^{2\dag}$, Kostas Daniilidis$^1$ \\[0ex]
$^1$ University of Pennsylvania \hspace{1em} $^2$ Zhejiang University\\ 
}

\twocolumn[{%
\renewcommand\twocolumn[1][]{#1}%
\maketitle
\begin{center}
    \newcommand{\teaserwidth}{\textwidth}
 \vspace{-0.25in}
    \centerline{
   \includegraphics[width=0.24\textwidth]{figures/final/ablative/abl_COCO_val2014_000000541055.jpg/img_cropped.jpg}
   \includegraphics[width=0.24\textwidth]{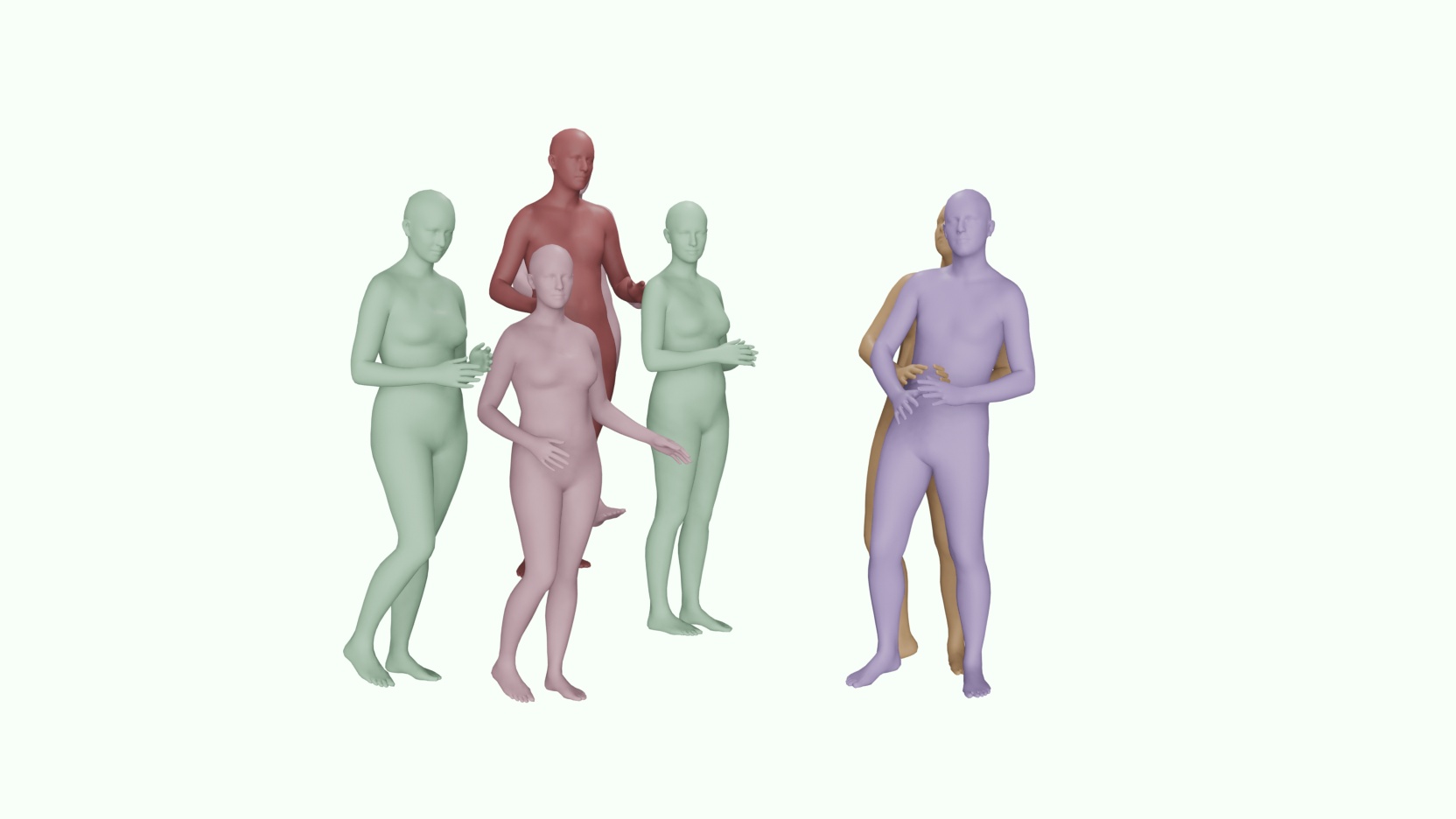}
   \includegraphics[width=0.12\textwidth]{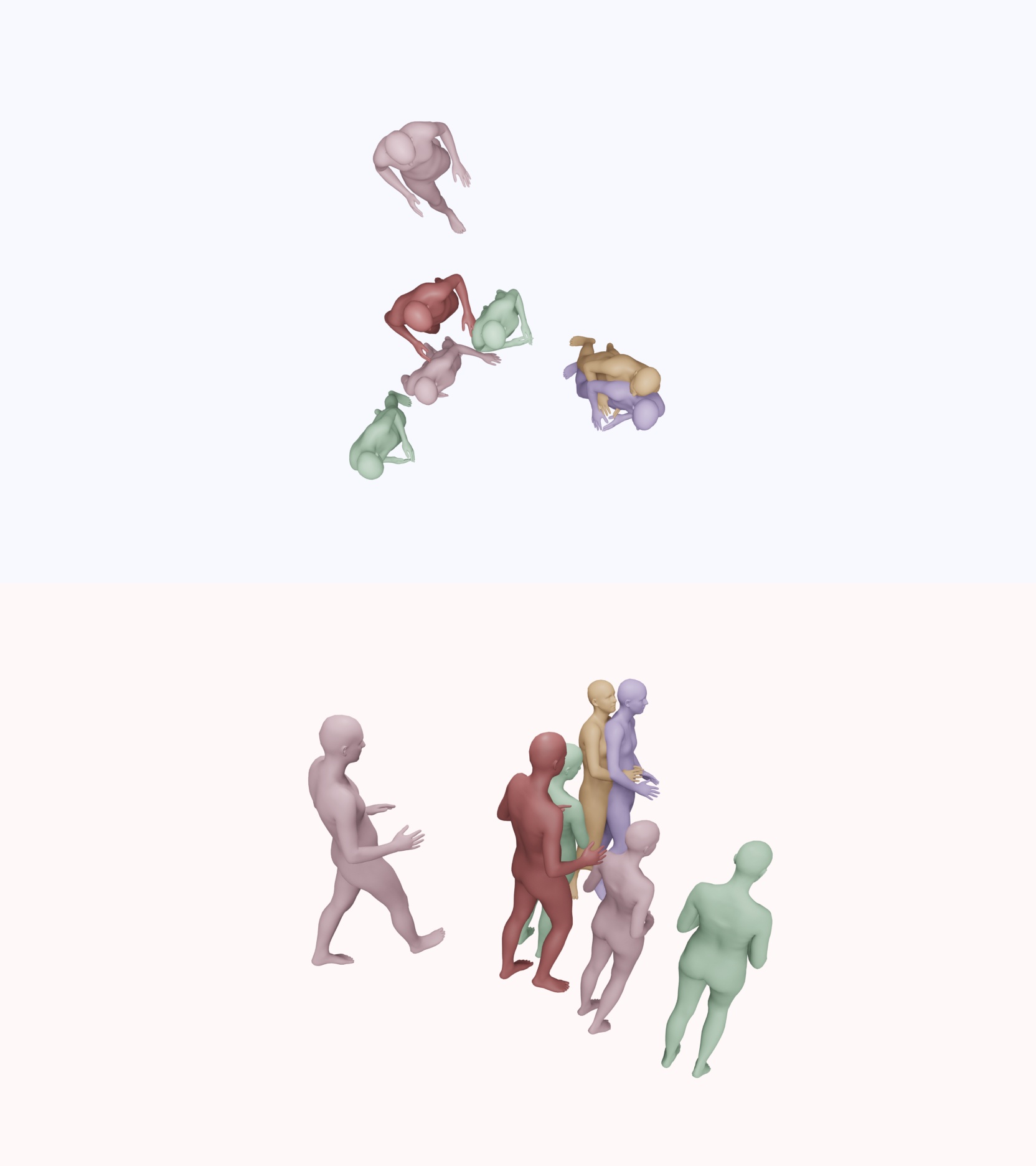}
   \includegraphics[width=0.24\textwidth]{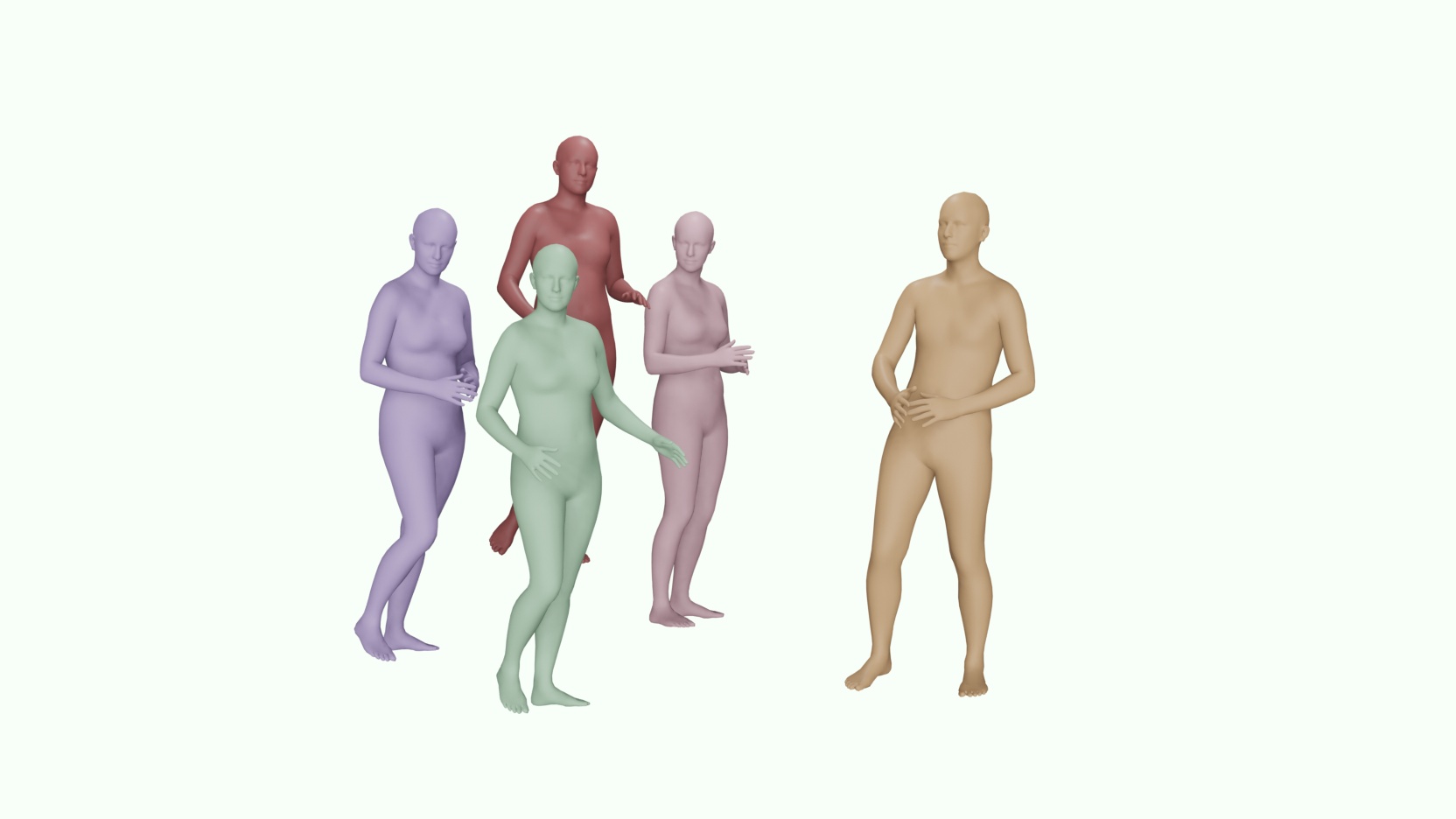}  
   \includegraphics[width=0.12\textwidth]{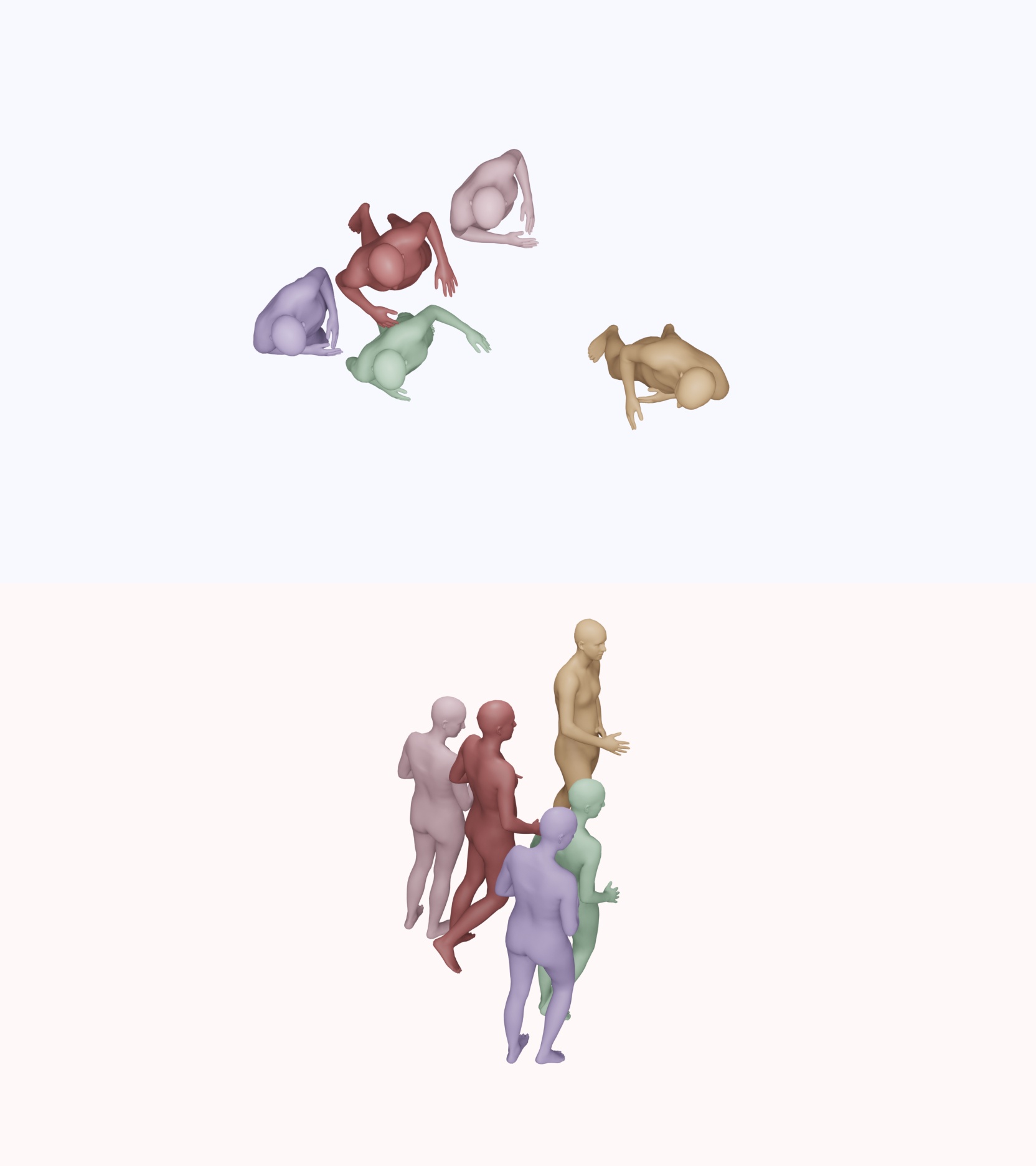}
 }
  \vspace{-1mm}
   \text{\scriptsize Input image } \hspace{40mm} \text{\scriptsize Baseline} \hspace{56mm} \text{\scriptsize Ours}  \hspace{12mm} \text{\color{white} \scriptsize}
	\vspace{-3mm}     
   \captionof{figure}{{\bf Coherent reconstruction of pose and shape for multiple people}. 
Typical top-down regression baselines (center) suffer from predicting people in overlapping positions, or in inconsistent depth orderings. Our approach (right) is trained to respect all these constraints and recover a coherent reconstruction of all the people in the scene in a feedforward manner.}
\label{fig:teaser}
\end{center}%
}]

\begin{abstract}
\vspace{-2mm}
In this work, we address the problem of multi-person 3D pose estimation from a single image. A typical regression approach in the top-down setting of this problem would first detect all humans and then reconstruct each one of them independently. However, this type of prediction suffers from incoherent results, e.g., interpenetration and inconsistent depth ordering between the people in the scene. Our goal is to train a single network that learns to avoid these problems and generate a coherent 3D reconstruction of all the humans in the scene. To this end, a key design choice is the incorporation of the SMPL parametric body model in our top-down framework, which enables the use of two novel losses. First, a distance field-based collision loss penalizes interpenetration among the reconstructed people. Second, a depth ordering-aware loss reasons about occlusions and promotes a depth ordering of people that leads to a rendering which is consistent with the annotated instance segmentation. This provides depth supervision signals to the network, even if the image has no explicit 3D annotations. The experiments show that our approach outperforms previous methods on standard 3D pose benchmarks, while our proposed losses enable more coherent reconstruction in natural images. The project website with videos, results, and code can be found at: {\footnotesize \url{https://jiangwenpl.github.io/multiperson}}
\vspace{-0.15in}
\blfootnote{$^*$ Equal contribution.}
\blfootnote{$^{\dag}$  X. Zhou and W. Jiang are affiliated with the State Key Lab of CAD\&CG, Zhejiang University. Email: xwzhou@zju.edu.cn.}   
\end{abstract}

\section{Introduction}
Recent work has achieved tremendous progress on the frontier of 3D human analysis tasks. Current approaches have established impressive performance for 3D keypoint estimation~\cite{martinez2017simple,sun2018integral}, 3D shape reconstruction~\cite{gabeur2019moulding,varol2018bodynet}, full-body 3D pose and shape recovery~\cite{guler2019holopose,kanazawa2018end,kolotouros2019learning,pavlakos2019texturepose}, or even going beyond that and estimating more detailed and expressive reconstructions~\cite{pavlakos2019expressive,xiang2019monocular}. However, as we progress towards more holistic understanding of scenes and people interacting in them, a crucial step is the coherent 3D reconstruction of multiple people from single images.

\begin{figure*}[!t]
	\centering
	\includegraphics[width=2.0\columnwidth,trim={0 450 190 0},clip]{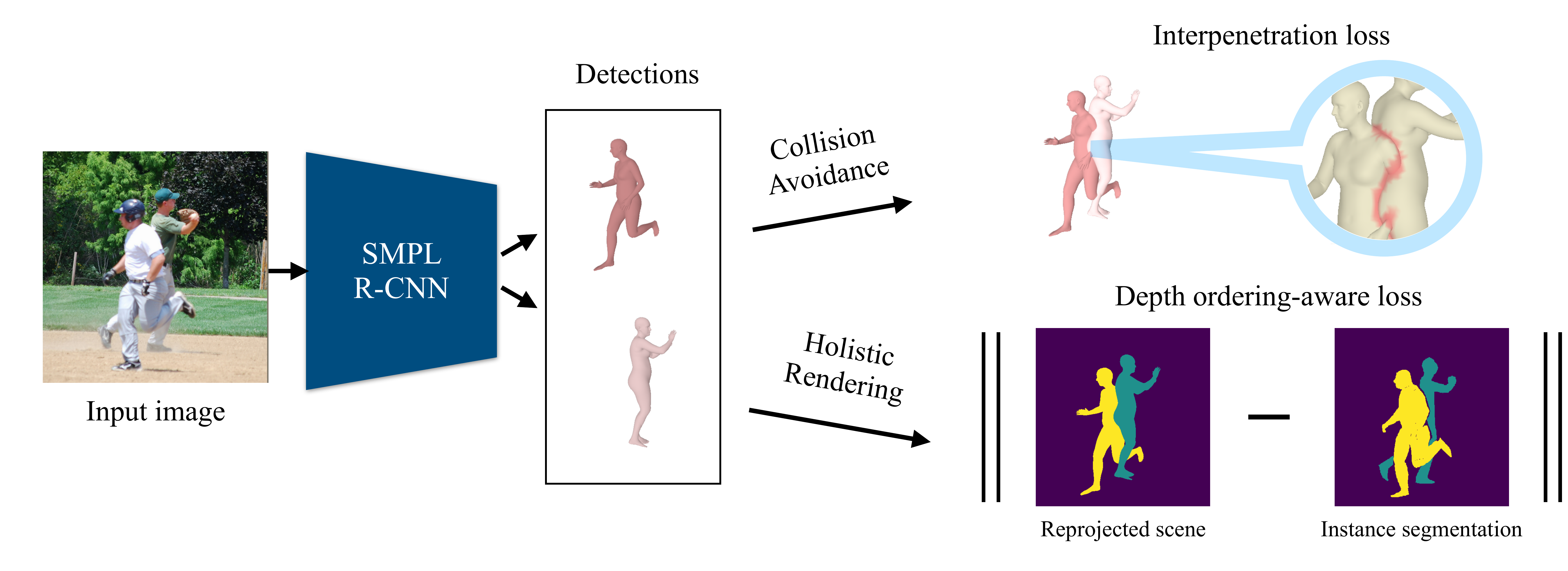}
	\vspace{-3mm}
	\caption{{\bf Overview of the proposed approach.}
We design an end-to-end framework for 3D pose and shape estimation of multiple people from a single image. An R-CNN-based architecture~\cite{he2017mask} detects all people in the image and estimates their SMPL parameters~\cite{loper2015smpl}. During training we incorporate constraints to promote a coherent reconstruction of all the people in the scene. First, we use an interpenetration loss to avoid people overlapping each other. Second, we apply a depth ordering-aware loss by rendering the meshes of all the people to the image and encouraging the rendered instance segmentation to match with the annotated instance masks.
}
\label{fig:pipeline}
\vspace{-5mm}
\end{figure*}

Regarding multi-person pose estimation, on one end of the spectrum, we have bottom-up approaches. The works following this paradigm, first detect all body joints in the scene and then group them, i.e., assigning them to the appropriate person. However, it is not straightforward how bottom-up processing can be extended beyond joints (e.g., use it for shape estimation, or mesh recovery). Different from bottom-up, top-down approaches first detect all people in the scene, and then estimate the pose for each one of them. Although they take a hard decision early on (person detection), they typically rely on state-of-the-art methods for person detection and pose estimation which allows them to achieve very compelling results, particularly in the 2D pose case, e.g.,~\cite{chen2018cascaded,sun2019deep,xiao2018simple}. However, when reasoning about the pose of multiple people in 3D, the problems can be more complicated than in 2D. For example, the reconstructed people can overlap each other in the 3D space, or be estimated at depths that are inconsistent with the actual depth ordering, as is demonstrated in Figure~\ref{fig:teaser}. This means that it is crucial to go beyond just predicting a reasonable 3D pose for each person individually, and instead estimate a coherent reconstruction of all the people in the scene.

This coherency of the holistic scene is the primary goal of this work. We adopt the typical top-down paradigm, and our aim is to train a deep network that learns to estimate a coherent reconstruction of all the people in the scene. Starting with a framework that follows the R-CNN pipeline~\cite{ren2015faster}, a key decision we make is to use of the SMPL parametric model~\cite{loper2015smpl} as our representation, and add a SMPL estimation branch to the R-CNN. The mesh representation provided by SMPL allows us to reason about occlusions and interpenetrations enabling the incorporation of two novel losses towards coherent 3D reconstruction. First, a common problem of predictions from regression networks is that the reconstructed people often overlap each other, since the feedforward nature does not allow for holistic feedback on the potential intersections. To train a network that learns to avoid this type of collisions, we introduce an interpenetration loss that penalizes intersections among the reconstructed people. This term requires no annotations and relies on a simple property of natural scenes, i.e., that people cannot intersect each other. Besides collisions, another source of incoherency in the results is that the estimated depths of the meshes are not respecting the actual depth ordering of the humans in the scene. Equipped with a mesh representation, we render our holistic scene prediction on the 2D image plane and penalize discrepancies of this rendering from the annotated instance segmentation. This loss enables reasoning about occlusion, encouraging the depth ordering of the people in the scene to be consistent with the annotated instance masks. Our complete framework (Figure~\ref{fig:pipeline}) is evaluated on various benchmarks and outperforms previous multi-person 3D pose and shape approaches, while the proposed losses improve coherency of the holistic result both qualitatively and quantitatively.

To summarize, our main contributions are:
\begin{itemize}
\item We present a complete framework for coherent regression of 3D pose and shape for multiple people.
\item We train with an interpenetration loss to avoid regressing meshes that intersect each other.
\item We train with a depth ordering-aware loss to promote reconstructions that respect the depth ordering of the people in the scene.
\item We outperfrom previous approaches for multi-person 3D pose and shape, while recovering significantly more coherent results.
\end{itemize}

\section{Related work}
In this Section we provide a short description of prior works that are more relevant to ours.

\textbf{Single-person 3D pose and shape}:
Many recent works estimate 3D pose in the form of a skeleton, e.g.,~\cite{martinez2017simple,mehta2017vnect,pavlakos2018ordinal,popa2017deep,sun2018integral,tekin2017learning,tome2017lifting,zhou2017towards}, or 3D shape in a non-parmetric way, e.g.,~\cite{gabeur2019moulding,smith2019facsimile,varol2018bodynet}. However, here we focus on full-body pose and shape reconstruction in the form of a mesh, typically using a parametric model, like SMPL~\cite{loper2015smpl}. After the early works on the problem~\cite{guan2009estimating,sigal2008combined}, the first fully automatic approach, SMPLify, was proposed by Bogo~\etal~\cite{bogo2016keep}. SMPLify iteratively fits SMPL on the 2D joints detected by a 2D pose estimation network~\cite{pishchulin2016deepcut}. This optimization approach was later extended in multiple ways; Lassner~\etal~\cite{lassner2017unite} use silhouettes for the fitting, Varol~\etal~\cite{varol2018bodynet} use voxel occupancy grids, while Pavlakos~\etal~\cite{pavlakos2019expressive} fit a more expressive parametric model, SMPL-X.

Despite the success of the aforementioned fitting approaches, recently we have observed an increased interest in approaches that regress the pose and shape parameters directly from images, using a deep network for this task. Many works focus on first estimating some form of intermediate representation before regressing SMPL parameters. Pavlakos~\etal~\cite{pavlakos2018learning} use keypoints and silhouettes, Omran~\etal~\cite{omran2018neural} use semantic part segmentation, Tung~\etal~\cite{tung2017self} append heatmaps for 2D joints to the RGB input, while Kolotouros~\etal~\cite{kolotouros2019convolutional} regress the mesh vertices with a Graph CNN. Regressing SMPL parameters directly from RGB input is more challenging, but it avoids any hand-designed bottleneck. Kanazawa~\etal~\cite{kanazawa2018end} use an adversarial prior to penalize improbable 3D shapes during training. Arnab~\etal~\cite{arnab2019exploiting} use temporal context to improve the regression network. G\"uler~\etal~\cite{guler2019holopose} incorporate a test-time post-processing based on 2D/3D keypoints and DensePose~\cite{guler2018densepose}.

\textbf{Multi-person 3D pose}:
For the multi-person case, the top-down paradigm is quite popular for 3D pose estimation, since it capitalizes on the success of the R-CNN works~\cite{girshick2015fast,ren2015faster,he2017mask}. The LCR-Net approaches~\cite{rogez2017lcr,rogez2019lcr} first detect each person, then classify its pose in a pose cluster and finally regress an offset for each joint. Dabral~\etal~\cite{dabral2019multi} first estimate 2D joints inside the bounding box and then regress 3D pose. Moon~\etal~\cite{moon2019camera} contribute a root network to give an estimate of the depth of the root joint. Zanfir~\etal~\cite{zanfir2018monocular} rely on scene constraints to iteratively optimize the 3D pose and shape of the people in the scene. Alternatively, there are also approaches that follow the bottom-up paradigm. Mehta~\etal~\cite{mehta2018single} propose a formulation based on occlusion-robust pose-maps, where Part Affinity Fields~\cite{cao2017realtime} are used for the association problem. Follow-up work~\cite{mehta2019xnect}, improves, among others, the robustness of the system. Finally, Zanfir~\etal~\cite{zanfir2018deep} solve a binary integer linear program to perform skeleton grouping.

In the context of pose and shape estimation in particular, there is a limited number of works that estimate full-body 3D pose and shape for multiple people in the scene. Zanfir~\etal~\cite{zanfir2018monocular} optimize the 3D shape of all the people in the image using multiple scene constraints. Our approach draws inspiration from this work and shares the same goal, in the sense of recovering a coherent 3D reconstruction. In contrast to them, instead of optimizing for this coherency at test-time, we train a feedforward regressor and use the scene constraints at training time to encourage it to produce coherent estimates at test-time. Using a feedforward network to estimate pose and shape for multiple people has been proposed by the work of Zanfir~\etal~\cite{zanfir2018deep}. However, in that case, 3D shape is regressed based on 3D joints, which are the output of a bottom-up system. In contrast, our approach is top-down, and SMPL parameters are regressed directly from pixels, instead of using an intermediate representation, like 3D joints. In fact, it is non-trivial to design a framework for SMPL parameter regression in a bottom-up manner.

\textbf{Coherency constraints}:
An important aspect of our work is the incorporation of loss terms that promote coherent 3D reconstruction of the multiple humans. Regarding our interpenetration loss, Bogo~\etal~\cite{bogo2016keep} and Pavlakos~\etal~\cite{pavlakos2019expressive} use a relevant objective to avoid self-interpenetrations of the human under consideration. In a more similar spirit to us, Zanfir~\etal~\cite{zanfir2018monocular} use a volume occupancy loss to avoid humans intersecting each other. In different applications, Hasson~\etal~\cite{hasson2019learning} penalize interpenetrations between the object and the hand that interacts with it, while Hassan~\etal~\cite{hassan2019resolving} penalize interpenetrations between humans and their environment. The majority of the above works uses the interpenetration penalty to iteratively refine estimates at test-time. With the exception of~\cite{hasson2019learning}, our work is the only one that uses an interpenetration term to guide the training of a feedforward regressor and promote colliding-free reconstructions at test time.

Regarding our depth ordering-aware loss, we follow the formulation of Chen~\etal~\cite{chen2016single}, which was also used in the  context of 3D human pose by Pavlakos~\etal~\cite{pavlakos2018ordinal}. In contrast to them, we do not use explicit depth annotations, but instead, we leverage the instance segmentation masks to reason about occlusion and thus, depth ordering. The work of Rhodin~\etal~\cite{rhodin2019neural} is also relevant, where inferring depth ordering is used as an intermediate abstraction for scene decomposition from multiple views. Our work also aims to estimate a coherent depth ordering, but we do so from a single image with the guidance of instance segmentation, while we retain a more explicit human representation in terms of meshes. Finally, using instance segmentation via render and compare has also been proposed by Kundu~\etal~\cite{kundu20183d}. However, their multi-instance evaluation includes only rigid objects, specifically cars, whereas we investigate the, significantly more complex, non-rigid case. 

\section{Technical approach}
In this Section, we describe the technical approach followed in this work. We start with providing some information about the SMPL model (Subsection~\ref{sec:smpl}) and the baseline regional architecture we use (Subsection~\ref{sec:baseline}). Then we describe in detail our proposed losses promoting interpenetration-free reconstruction (Subsection~\ref{sec:interpenetration}) and consistent depth ordering (Subsection~\ref{sec:depth}). Finally, we provide more implementation details (Subsection~\ref{sec:details}).

\subsection{SMPL parametric model} \label{sec:smpl}
For the human body representation, we use the SMPL parametric model of the human body~\cite{loper2015smpl}. What makes SMPL very appropriate for our work, in comparison with other representations, is that it allows us to reason about occlusion and interpenetration enabling the use of the novel losses we incorporate in the training of our network. The SMPL model defines a function $\mathcal{M}(\pose, \shape)$ that takes as input the pose parameters $\pose$, and the shape parameters $\shape$, and outputs a mesh $M \in \mathbb{R}^{N_v \times 3}$, consisting of $N_v = 6890$ vertices. The model also offers a convenient mapping from mesh vertices to $k$ body joints $J$, through a linear regressor $W$, such that joints can be expressed as a linear combination of mesh vertices, $J = WM$.

\subsection{Baseline architecture} \label{sec:baseline}
In terms of the architecture for our approach, we follow the familiar R-CNN framework~\cite{ren2015faster}, and use a structure that is most similar to the Mask R-CNN iteration~\cite{he2017mask}. Our network consists of a backbone (here ResNet50~\cite{he2016deep}), a Region Proposal Network, as well as heads for detection and SMPL parameter regression (SMPL branch). Regarding the SMPL branch, its architecture is similar to the iterative regressor proposed by Kanazawa~\etal~\cite{kanazawa2018end}, regressing pose and shape parameters, $\pose$ and $\shape$ respectively, as well as camera parameters $\cam = \{s, t_x, t_y\}$. The camera parameters are predicted per bounding box but we later update them based on the position of the bounding box in the full image (details in the Sup.Mat.). Although there is no explicit feedback among bbox predictions, the receptive field of each proposal includes the majority of the scene. Since each bounding box is aware of neighboring people and their poses, it can make an {\it informed pose prediction} that is consistent with them.

For our baseline network, the various components are trained jointly in an end-to-end manner. The detection task is trained according to the training procedure of~\cite{he2017mask}, while for the SMPL branch, the training details are similar to the ones proposed by Kanazawa~\etal~\cite{kanazawa2018end}. In the rare cases that 3D ground truth is available, we apply a loss, $L_{3D}$, on the SMPL parameters and the 3D keypoints. In the most typical case that only 2D joints are available, we use a 2D reprojection loss, $L_{2D}$, to minimize the distance between the ground truth 2D keypoints and the projection of the 3D joints, $J$, to the image. Additionally, we also use a discriminator and apply an adversarial prior $L_{adv}$ on regressed pose and shape parameters, to encourage the output bodies to lie on the manifold of human bodies. Each of the above losses is applied independently to each proposal, after assigning it to the corresponding ground truth bounding box. More details about the above loss terms and the training of the baseline model are included in the Sup.Mat.

\begin{figure}[!t]
	\centering
	\includegraphics[scale=0.07,trim={0 70 0 0},clip]{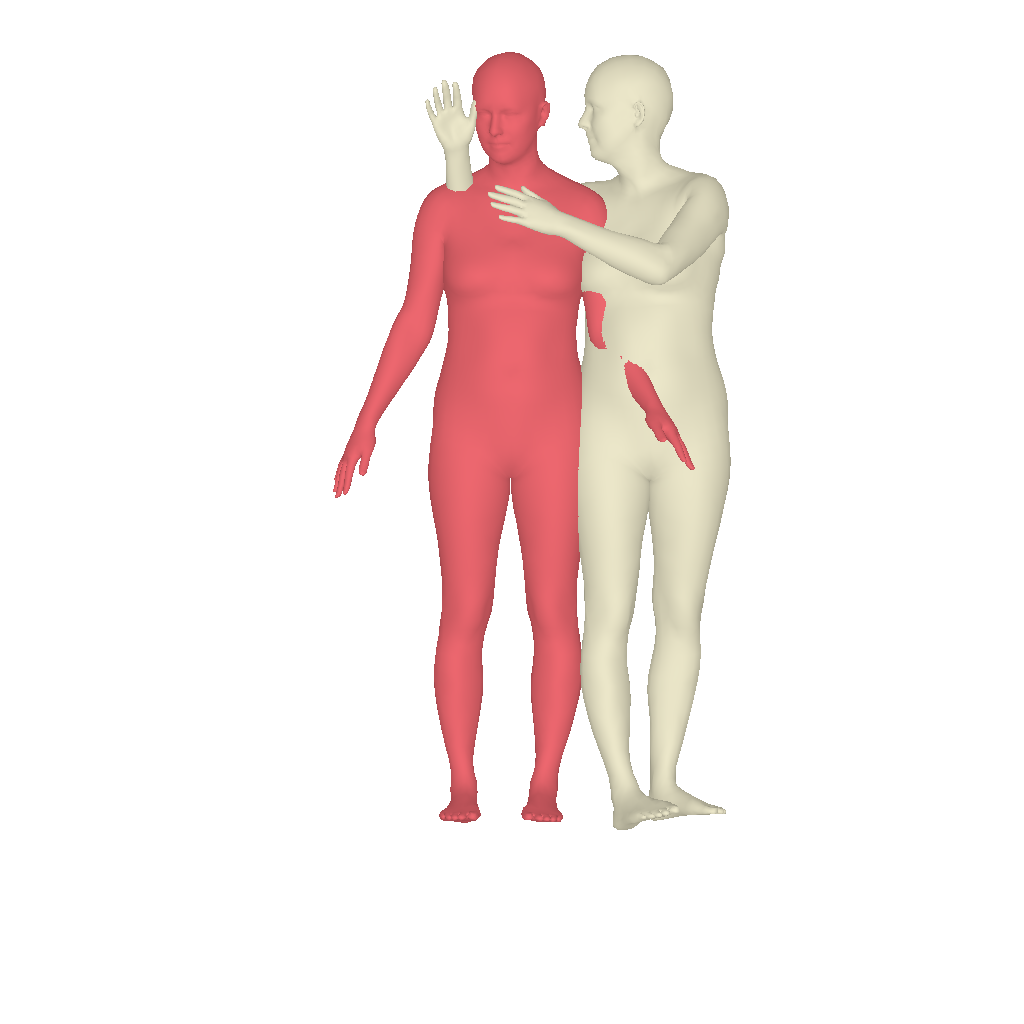}
	\includegraphics[scale=0.07,trim={0 0 0 0},clip]{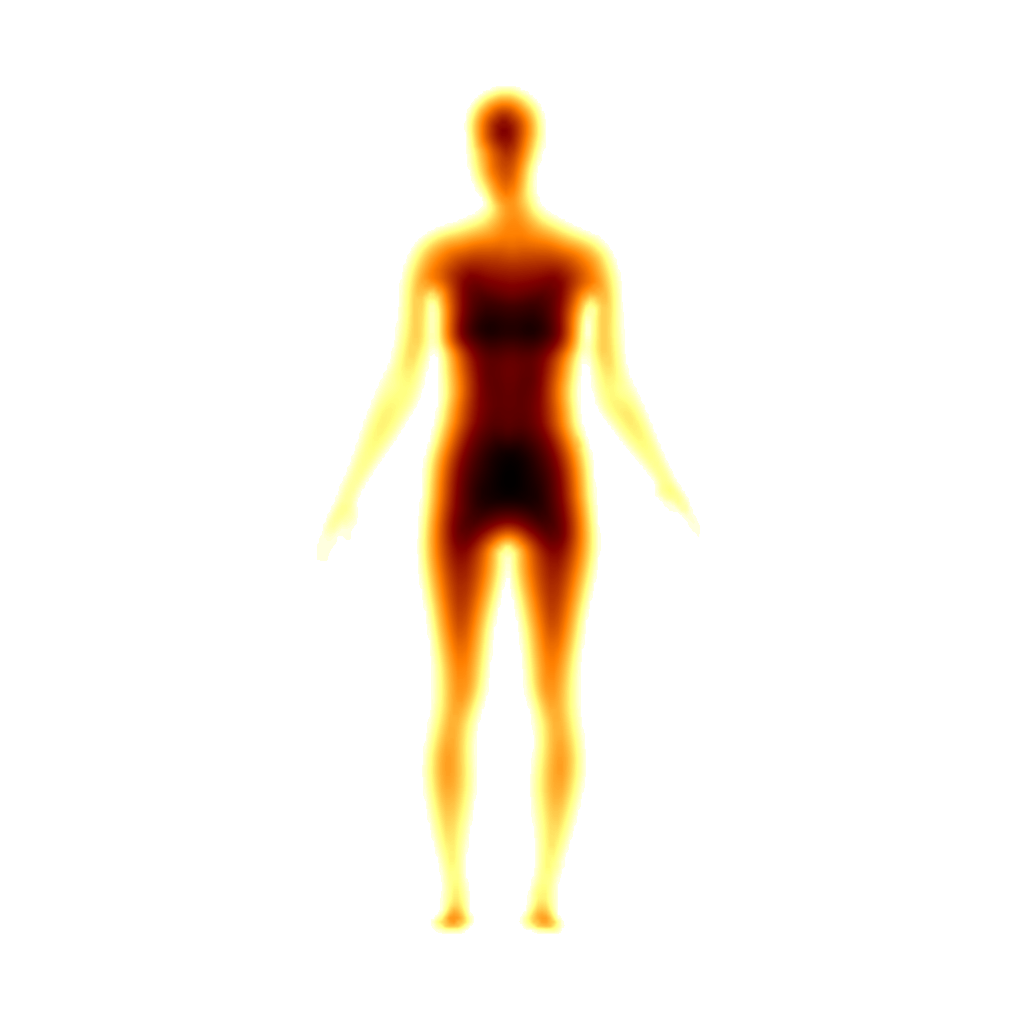}
	\includegraphics[scale=0.07,trim={0 70 0 0},clip]{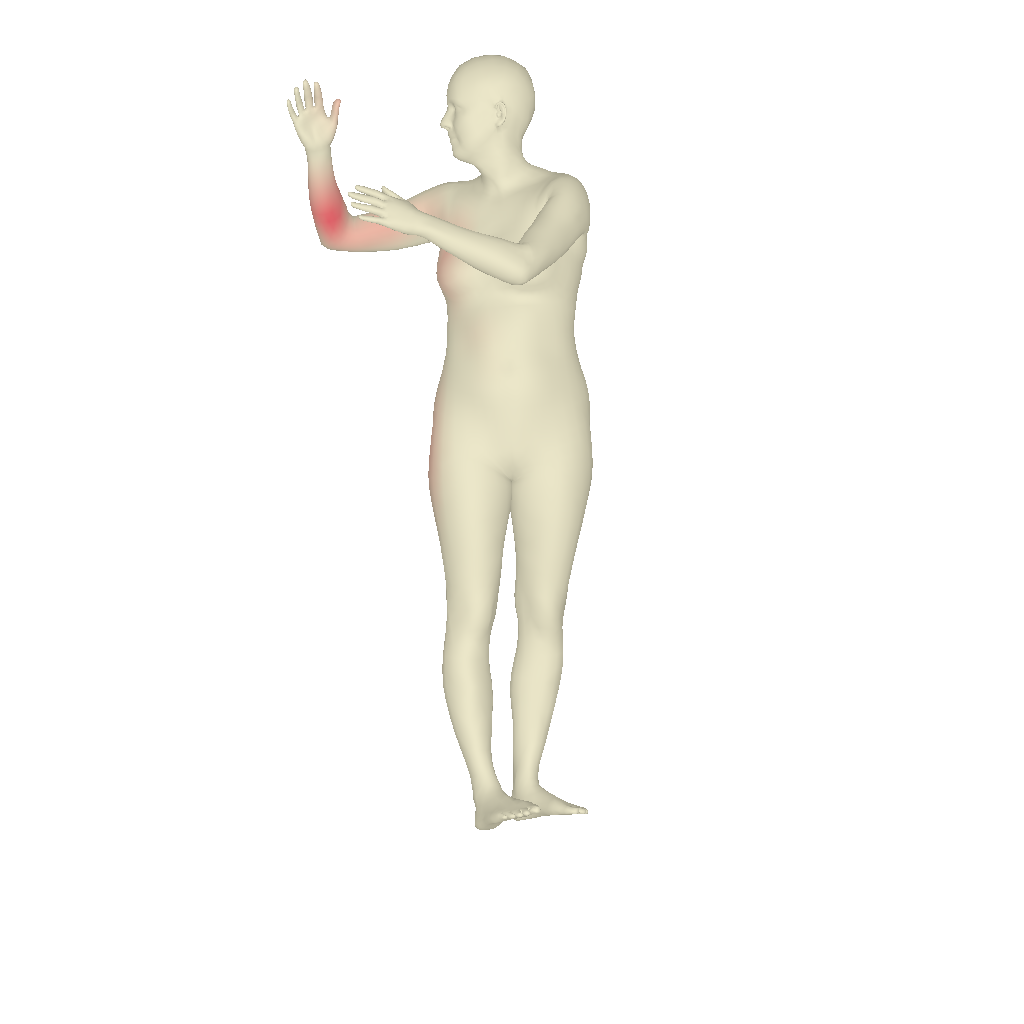}	
	\vspace{-2mm}
	\caption{{\bf Illustration of interpenetration loss.}
Left: Collision between person $i$ (red) and $j$ (beige). Center: Distance field $\phi_i$ for person $i$, Right: Mesh $M_j$ of person $j$. The vertices of $M_j$ that collide with person $i$, i.e., located in non-zero areas of $\phi_i$ and visualized with soft red, are penalized by the interpenetration loss.
}
\label{fig:interpenetration}
\vspace{-2mm}
\end{figure}

\subsection{Interpenetration loss} \label{sec:interpenetration}
A critical barrier towards coherent reconstruction of multiple people from a single image is that the regression network can often predict the people to be in overlapping locations. To promote prediction of non-colliding people, we introduce a loss that penalizes interpenetrations among the reconstructed people. Our formulation draws inspiration from~\cite{hassan2019resolving}.  An important difference is that instead of a static scene and a single person, our scene includes multiple people and it is generated in a dynamic way during training.

\begin{figure*}[!t]
	\centering
	\includegraphics[scale=0.19]{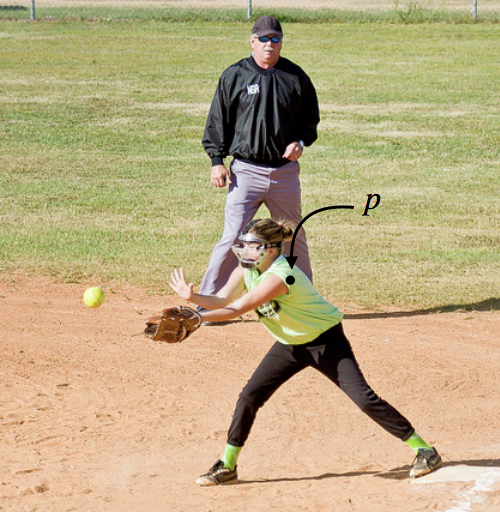}
	\includegraphics[scale=0.19]{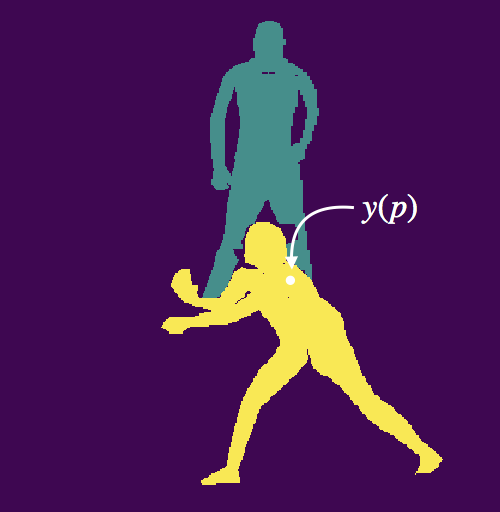}
	\includegraphics[scale=0.19]{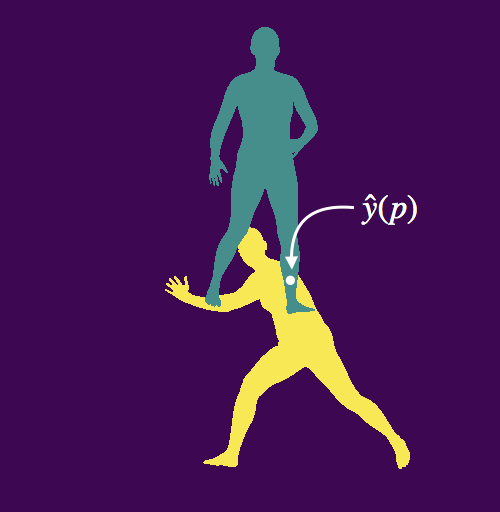}
	\includegraphics[scale=0.19]{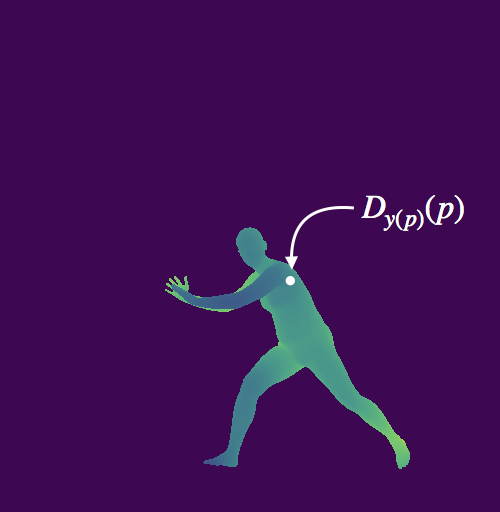}
	\includegraphics[scale=0.19]{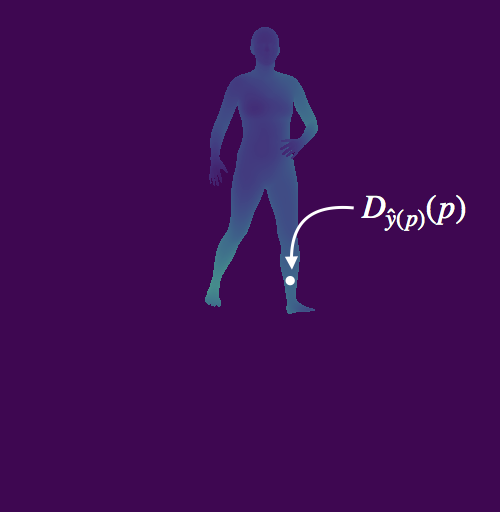}
	\vspace{-2mm}
	\caption{{\bf Illustration of depth ordering-aware loss.}
For an RGB image (first image), we consider the annotated instance segmentation (second image), and the instances based on the rendering of the estimated meshes on the image plane (third image). In case that there is a disagreement between the person index, e.g., for pixel $p$, where $y(p) \neq \hat{y}(p)$, we penalize the corresponding depth estimates at this pixel with an ordinal depth loss. The pixel depths $D_{y(p)}(p)$ and $D_{\hat{y}(p)}(p)$ are estimated by rendering the depth map independently for each person mesh (fourth and fifth image). This allows gradients to be backpropagated even to the non-visible vertices.
}
\label{fig:depth}
\vspace{-2mm}
\end{figure*}

Let $\phi$ be a modified Signed Distance Field (SDF) for the scene that is defined as follows:
\begin{equation}
\phi(x,y,z) = -\min\left(\operatorname{SDF}(x,y,z), 0\right),
\end{equation}
According to the above definition, inside each human, $\phi$ takes positive values, proportional to the distance from the surface, while it is simply $0$ outside of the human. Typically, $\phi$ is defined on a voxel grid of dimensions $N_p \times N_p \times N_p$. The na\"ive generation of a single voxelized representation for the whole scene is definitely possible. However, we often require a very fine voxel grid, which depending on the extend of the scene, might make processing intractable in terms of memory and computation. One critical observation here is that we can compute a separate $\phi_i$ function for each person in the scene, by calculating a tight box around the person and voxelizing it. This allows us to ignore empty scene space that is not covered by any person and we can instead use a fine spatial resolution to get a detailed voxelization of the body. Using this formulation, the collision penalty of person $j$ for colliding with person $i$ is defined as:
\begin{equation}
\mathcal{P}_{ij} = \sum_{v \in M_j} \tilde{\phi}_i(v),
\end{equation}
where $\tilde{\phi}_i(v)$ samples the $\phi_i$ value for each 3D vertex $v$ in a differentiable way from the 3D grid using trilinear interpolation (Figure~\ref{fig:interpenetration}). The $\phi_i$ computation for person $i$ is performed by a custom GPU implementation. This computation does not have to be differentiable; $\phi_i$ only defines a distance field from which we sample values in a differentiable way. By definition, $\mathcal{P}_{ij}$ is non-negative. It takes value $0$ if there is no collision between person $i$ and $j$ and increases as the distance of the surface vertices for person $j$ move farther from the surface of person $i$. In theory, $\mathcal{P}_{ij}$ can be used by itself as an optimization objective for interpenetration avoidance. However, in practice, we observed that it results in very large gradients for the person translation, leading to training instabilities when there are heavy collisions. Instead of the typical term, we use a robust version of this objective. More specifically, our final interpenetration loss for a scene with $N$ people is defined as follows:
\begin{equation}
L_{\mathcal{P}} = \sum_{j=1}^N \rho \left( \sum_{i=1, i \neq j}^N \mathcal{P}_{ij}\right)
\end{equation}
where $\rho$ is the Geman-McClure robust error function~\cite{geman1987statistical}. To avoid penalizing intersections between boxes corresponding to the same person, we use only the most confidence box proposal assigned to a ground truth box. 

\subsection{Depth ordering-aware loss} \label{sec:depth}
Besides interpenetration, another common problem in multi-person 3D reconstruction is that people are often estimated in incorrect depth order. This problem is more evident in cases where people overlap on the 2D image plane. Although it is obvious to the human eye which person is closer (due to the occlusion), the network predictions can still be incoherent. Fixing this depth ordering problem would be easy if we had access to pixel-level depth annotations. However, this type of annotations is rarely available. Our key idea here is that we can leverage the instance segmentation annotations that are often available, e.g., in the large scale COCO dataset~\cite{lin2014microsoft}. Rendering the meshes of all the reconstructed people on the image plane can indicate the person corresponding to each pixel and optimize based on the agreement with the annotated instance annotation.

Although this idea sounds straightforward, its realization is more complicated. An obvious implementation would be to use a differentiable renderer, e.g., the Neural Mesh Renderer (NMR)~\cite{kato2018neural}, and penalize inconsistencies between the actual instance segmentation and the one produced by rendering the meshes to the image. The practical problem with~\cite{kato2018neural} is that it backpropagates errors only to visible mesh vertices; if there is a depth ordering error, it will not promote the invisible vertices to move closer to the camera. In practice, we observed that this tends to move most people farther away, collapsing our training. Liu~\etal~\cite{liu2019soft} attempt to address this problem, but we observed that their softmax operation across the depths can result in vanishing gradients, while we also faced numerical instabilities.

Instead of rendering only the semantic segmentation of the scene, we also render the depth image $D_i$ for each person independently using NMR~\cite{kato2018neural}. Assuming the scene has $N$ people, we assign a unique index $i \in \{1,2,\dots,N\}$ to each one of them. Let $y(p)$ be the person index at pixel location $p$ in the ground truth segmentation, and $\hat{y}(p)$ be the predicted person index based on the rendering of the 3D meshes. We use $0$ to indicate background pixels. If for a pixel $p$ the two estimates indicate a person (no background) and disagree, i.e., $y(p) \neq \hat{y}(p)$, then we apply a loss to the depth values of both people for this pixel, $y(p)$ and $\hat{y}(p)$, to promote the correct depth ordering. The loss we apply is an ordinal depth loss, similar in spirit to~\cite{chen2016single}. More specifically, the complete loss expression is:
\begin{equation}
L_{\mathcal{D}} = \sum_{p \in \mathcal{S}} \log\left(1+\exp\left(D_{{y}(p)}(p) - D_{\hat{y}(p)}(p)\right)\right)
\end{equation}
where $\mathcal{S} = \{p \in I:\, y(p) > 0,  \hat{y}(p) > 0,y(p) \neq \hat{y}(p)\}$ represents the set of pixels for image $I$ where we have depth ordering mistakes  (Figure~\ref{fig:depth}). The key detail here is that the loss is backpropagated to the mesh (and eventually the model parameters) of both people, instead of backpropagating gradients only to the visible person, as a conventional differentiable renderer would do. This promotes a more symmetric nature to the loss (and the updates), and eventually makes this loss practical.

\subsection{Implementation details} \label{sec:details}
Our implementation is done using PyTorch and the publicly available mmdetection library~\cite{mmdetection}. We resize all input images to 512x832, keeping the same aspect ratio as in the original COCO training. For the baseline model we train only with the losses specified in Subsection~\ref{sec:baseline}, while for our full model we include in our training the losses proposed in Subsections~\ref{sec:interpenetration} and~\ref{sec:depth}. Our training uses 2 1080Ti GPUs and a batch size of 4 images per GPU.

For the SDF computation, we reimplemented~\cite{stutz2017learning,stutz2018learning} in CUDA. Voxelizing a single mesh in a $32\times 32 \times 32$ voxel grid requires about 45ms on an 1080Ti GPU. For efficiency, we perform 3D bounding box checks to detect overlapping 3D bounding boxes, and voxelize only the relevant meshes. Additionally, we reimplemented parts of NMR~\cite{kato2018neural} to make rendering large images more efficient. This allowed us to have more than an order of magnitude of speedup since the forward pass complexity dropped from $O(Fwh)$ to $O(F + wh)$ on average, where $F$ is the number of faces and $w$ and $h$ the image width and height respectively.

\section{Experiments}
In this Section, we present the empirical evaluation of our approach. First, we describe the datasets used for training and evaluation (Subsection~\ref{sec:datasets}). Then, we focus on the quantitative evaluation (Subsections~\ref{sec:sota} and~\ref{sec:ablatives}), and finally we present more qualitative results (Subsection~\ref{sec:qualitative}).

\subsection{Datasets} \label{sec:datasets}

\noindent
\textbf{Human3.6M}~\cite{ionescu2013human3}:
It is an indoor dataset where a single person is visible in each frame. It provides 3D ground truth for training and evaluation. We use Protocol 2 of~\cite{kanazawa2018end}, where Subjects S1,S5,S6,S7 and S8 are used for training, while Subjects S9 and S11 are used for evaluation.

\noindent
\textbf{MuPoTS-3D}~\cite{mehta2018single}:
It is a multi-person dataset providing 3D ground truth for all the people in the scene. We use this dataset for evaluation using the same protocol as~\cite{mehta2018single}.

\noindent
\textbf{Panoptic}~\cite{joo2015panoptic}:
It is a dataset with multiple people captured in the Panoptic studio. We use this dataset for evaluation, following the protocol of~\cite{zanfir2018monocular}.

\noindent
\textbf{MPI-INF-3DHP}~\cite{mehta2017monocular}:
It is a single person dataset with 3D pose ground truth. We use subjects S1 to S8 for training.

\noindent
\textbf{PoseTrack}~\cite{andriluka2018posetrack}:
In-the-wild dataset with 2D pose annotations. Includes multiple frames for each sequence. We use this dataset for training and evaluation.

\noindent
\textbf{LSP}~\cite{johnson2010clustered}, \textbf{LSP Extended}~\cite{johnson2011learning}, \textbf{MPII}~\cite{andriluka20142d}:
In-the-wild datasets with annotations for 2D joints. We use the training sets of these datasets for training.

\noindent
\textbf{COCO}~\cite{lin2014microsoft}:
In-the-wild dataset with 2D pose and instance segmentation annotations. We use the 2D joints for training as we do with the other in the-wild datasets, while the instance segmentation masks are employed for the computation of the depth ordering-aware loss.

\subsection{Comparison with the state-of-the-art} \label{sec:sota}
For the comparison with the state-of-the-art, as a sanity check, we first evaluate the performance of our approach on a typical single person baseline. Our goal is always multi-person 3D pose and shape, but we expect our approach to achieve competitive results, even in easier settings, i.e., when only one person is in the image. More specifically, we evaluate the performance of our network on the popular Human3.6M dataset~\cite{ionescu2013human3}. The most relevant approach here is HMR by Kanazawa~\etal~\cite{kanazawa2018end}, since we share similar architectural choices (iterative regressor, regression target), training practices (adversarial prior) and training data. The results are presented in Table~\ref{table:h36m}. Our approach outperforms HMR, as well as the approach of Arnab~\etal~\cite{arnab2019exploiting}, that uses the same network with HMR, but is trained with more data.

\begin{table}\centering
	\begin{tabular}{c|c|c|c}
		\toprule
		Method & HMR~\cite{kanazawa2018end} & Arnab~\etal~\cite{arnab2019exploiting} & Ours  \\
		\midrule
		Reconst. Error & 56.8 & 54.3 & \textbf{52.7}\\
		\bottomrule
	\end{tabular}
	\caption{\textbf{Results on Human3.6M}. The numbers are mean 3D joint errors in mm after Procrustes alignment (Protocol 2). The results of all approaches are obtained from the original papers.}
	\label{table:h36m}
\end{table}

Having established that our approach is competitive in the single person setting, we continue the evaluation with multi-person baselines. In this case, we consider approaches that also estimate pose and shape for multiple people. The most relevant baselines are the works of Zanfir~\etal~\cite{zanfir2018monocular,zanfir2018deep}. We compare with these approaches in the Panoptic dataset~\cite{joo2015panoptic,joo2017panoptic}, using their evaluation protocol (assuming no data from the Panoptic studio are used for training). The full results are reported in Table~\ref{table:panoptic}. Our initial network (baseline), trained without our proposed losses, achieves performance comparable with the results reported by the previous works of Zanfir~\etal. More importantly though, adding the two proposed losses (full), improves performance across all subsequences and overall, while we also outperform the previous baselines. These results demonstrate both the strong performance of our approach in the multi-person setting, as well as the benefit we get from the losses we propose in this work.

\begin{table}\centering
\footnotesize
	\begin{tabular}{c|c|c|c|c|c}
	\toprule
	Method & Haggling & Mafia & Ultim. & Pizza & Mean\\
	\midrule
	Zanfir~\etal~\cite{zanfir2018monocular} & 140.0 & 165.9 & 150.7 & \textbf{156.0} & 153.4\\ 
	Zanfir~\etal~\cite{zanfir2018deep} & 141.4 & 152.3 & \textbf{145.0} & 162.5 & 150.3\\ 	
	Ours (baseline) & 141.2 & 140.3 & 160.7 & 156.8 & 149.8\\	
	Ours (full) & \textbf{129.6} & \textbf{133.5} & 153.0 & 156.7 & \textbf{143.2}\\
	\bottomrule
	\end{tabular}
	\caption{\textbf{Results on the Panoptic dataset}. The numbers are mean per joint position errors after centering the root joint. The results of all approaches are obtained from the original papers.}
	\vspace{-2mm}
	\label{table:panoptic}
\end{table}

Another popular benchmark for multi-person 3D pose estimation is the MuPoTS-3D dataset~\cite{mehta2017monocular}. Since no multi-person 3D pose and shape approach reports results on this benchmark, we implement two strong top-down baselines, based on state-of-the-art approaches for single-person 3D pose and shape. Specifically, we select a regression approach, HMR~\cite{kanazawa2018end}, and an optimization approach, SMPLify-X~\cite{pavlakos2019expressive}, and we apply them on detections provided by OpenPose~\cite{cao2019openpose} (as is suggested by their public repositories), or by Mask-RCNN~\cite{he2017mask} (for the case of HMR). The full results are reported in Table~\ref{table:mupots}. As we can see, our baseline model performs comparably to the other approaches, while our full model trained with the proposed losses improves significantly over the baseline. Similarly with the previous results, this experiment further justifies the use of our coherency losses. Besides this, we also demonstrate that na\"ive baselines trained with a single person in mind are suboptimal for the multi-person setting of 3D pose. This is different from the 2D case, where a single-person network can perform particularly well in multi-person top-down pipelines as well, e.g.,~\cite{chen2018cascaded,sun2019deep,xiao2018simple}. For the 3D case though, when multiple people are involved, making the network aware of occlusions and interpenetrations during training, can actually be beneficial at test-time too.

\begin{table}\centering
	\begin{tabular}{c|c|c}
	\toprule
	Method & All & Matched\\
	\midrule	
	OpenPose + SMPLify-X~\cite{pavlakos2019expressive} & 62.84 & 68.04\\
	OpenPose + HMR~\cite{kanazawa2018end} & 66.09 & 70.90\\	
	Mask-RCNN + HMR~\cite{kanazawa2018end} & 65.57 & 68.57\\			
	Ours (baseline) & 66.95 & 68.96\\	
	Ours (full) & \textbf{69.12} & \textbf{72.22}\\
	\bottomrule
	\end{tabular}
	\caption{\textbf{Results on MuPoTS-3D}. The numbers are 3DPCK. We report the overall accuracy (All), and the accuracy only for person annotations matched to a prediction (Matched).}
	\label{table:mupots}
\end{table}

\subsection{Ablative studies} \label{sec:ablatives}
For this work, our interest in multi-person 3D pose estimation extends beyond just estimating poses that are accurate under the typical 3D pose metrics. Our goal is also to recover a coherent reconstruction of the scene. This is important, because in many cases we can improve the 3D pose metrics, e.g., get a better 3D pose for each detected person, but return incoherent results holistically. For example, the depth ordering of the people might be incorrect, or the reconstructed meshes might be positioned such that they overlap each other. To demonstrate how our proposed losses improve the network predictions under these coherency metrics even if they are only applied during training, we perform two ablative studies for more detailed evaluation.

First, we expect our interpenetration loss to naturally eliminate most of the overlapping people in our predictions. We evaluate this on MuPoTS-3D and PoseTrack, reporting the number of collisions with and without the interpenetration loss. The results are reported in Table~\ref{table:interpenetration}. As we expected, we observe significant decrease in the number of collisions when we train the network with the $L_{\mathcal{P}}$ loss.

\begin{table}\centering
	\begin{tabular}{c|c|c}
		\toprule
		Method & MuPoTS-3D & PoseTrack \\
		\midrule
	    Our baseline & 114 & 653 \\
	    Our baseline + $L_{\mathcal{P}}$  & 34 & 202 \\
		\bottomrule
	\end{tabular}
	\caption{\textbf{Ablative for interpenetration loss}. The results indicate the number of collisions on MuPoTS-3D and PoseTrack.}
	\vspace{-2mm}
	\label{table:interpenetration}
\end{table}

Moreover, our depth ordering-aware loss should improve the translation estimates for the people in the scene. Since for monocular methods it is not meaningful to evaluate metric translation estimates, we propose to evaluate only the returned depth ordering. More specifically, we consider all pairs of people in the scene, and we evaluate whether our method predicted the ordinal depth relation for this pair correctly. In the end, we report the percentage of correctly estimated ordinal relations in Table~\ref{table:ordering}. As expected, the depth ordering-aware loss improves upon our baseline. In the same Table, we also report the results of the approach of Moon~\etal~\cite{moon2019camera} which is the state-of-the-art for 3D skeleton regression. Although~\cite{moon2019camera} is skeleton-based and thus, not directly comparable to us, we want to highlight that even a state-of-the-art approach (under 3D pose metric evaluation) can still suffer from incoherency in the results. This provides evidence that we often might overlook the coherency of the holistic reconstruction, and we should also consider this aspect when we evaluate the quality of our results.

Finally, we underline that we do not apply these coherency losses at test time. Instead, during training, our losses act as constraints to the reconstruction and ultimately provide better supervision to the network, for images that no explicit 3D annotations are available. The improved supervision leads to more coherent results {\it at test time too}.

\begin{table}\centering
\footnotesize
	\begin{tabular}{c|c|c|c}
	\toprule
	Method & Moon~\etal~\cite{moon2019camera} & Our baseline & Our baseline + $L_{\mathcal{D}}$ \\
	\midrule
	Accuracy & 90.85\% & 92.17\% & 93.68\%\\
	\bottomrule
	\end{tabular}
	\caption{\textbf{Ablative for depth-ordering-aware loss.} Depth ordering results on MuPoTS-3D. We consider all pairs of people in the image, and we evaluate whether the approaches recovered the ordinal depth relation between the two people correctly. The numbers are percentages of correctly estimated ordinal depth relations. }
	\vspace{-2mm}
	\label{table:ordering}
\end{table}

\begin{figure*}[!t]
	\centering
    \includegraphics[width=0.24\textwidth]{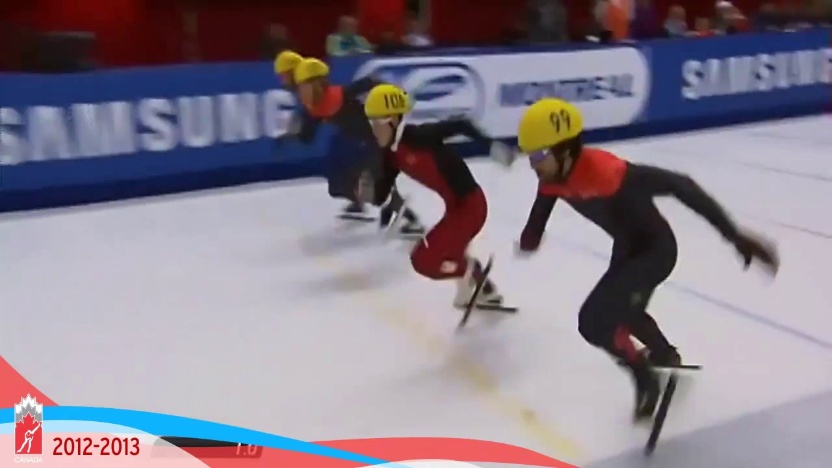}
    \includegraphics[width=0.24\textwidth]{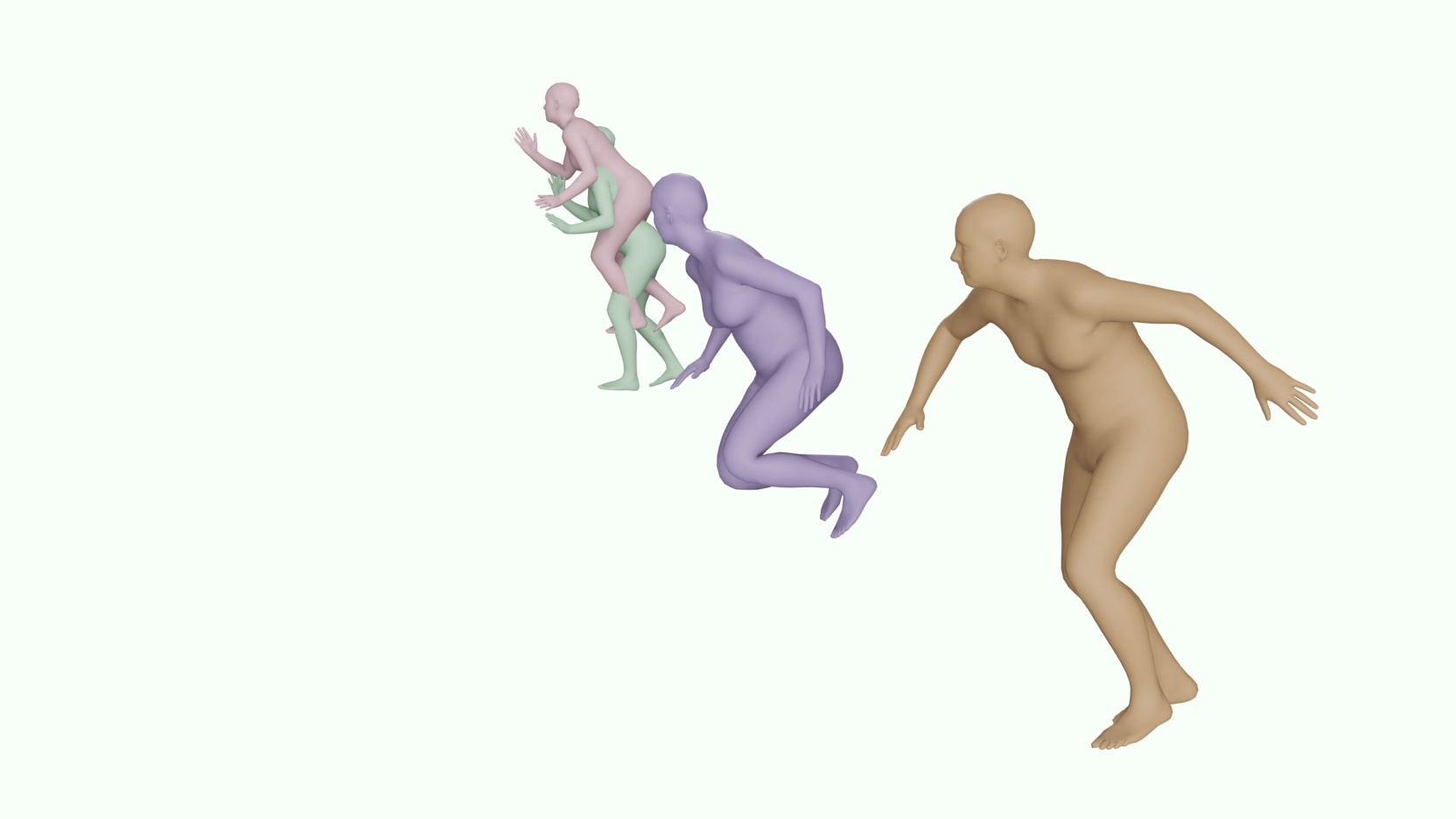}
    \includegraphics[width=0.12\textwidth]{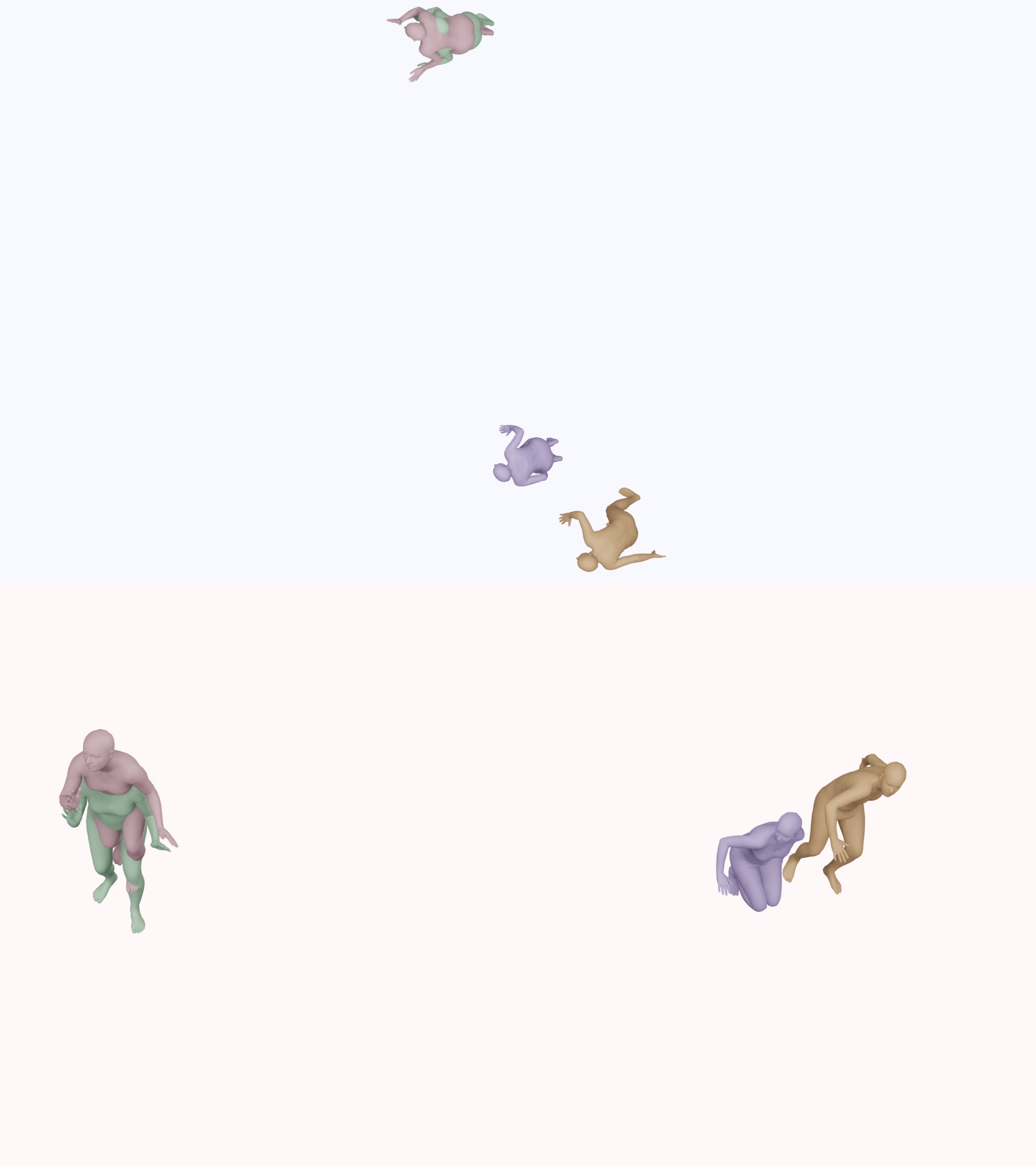}
    \includegraphics[width=0.24\textwidth]{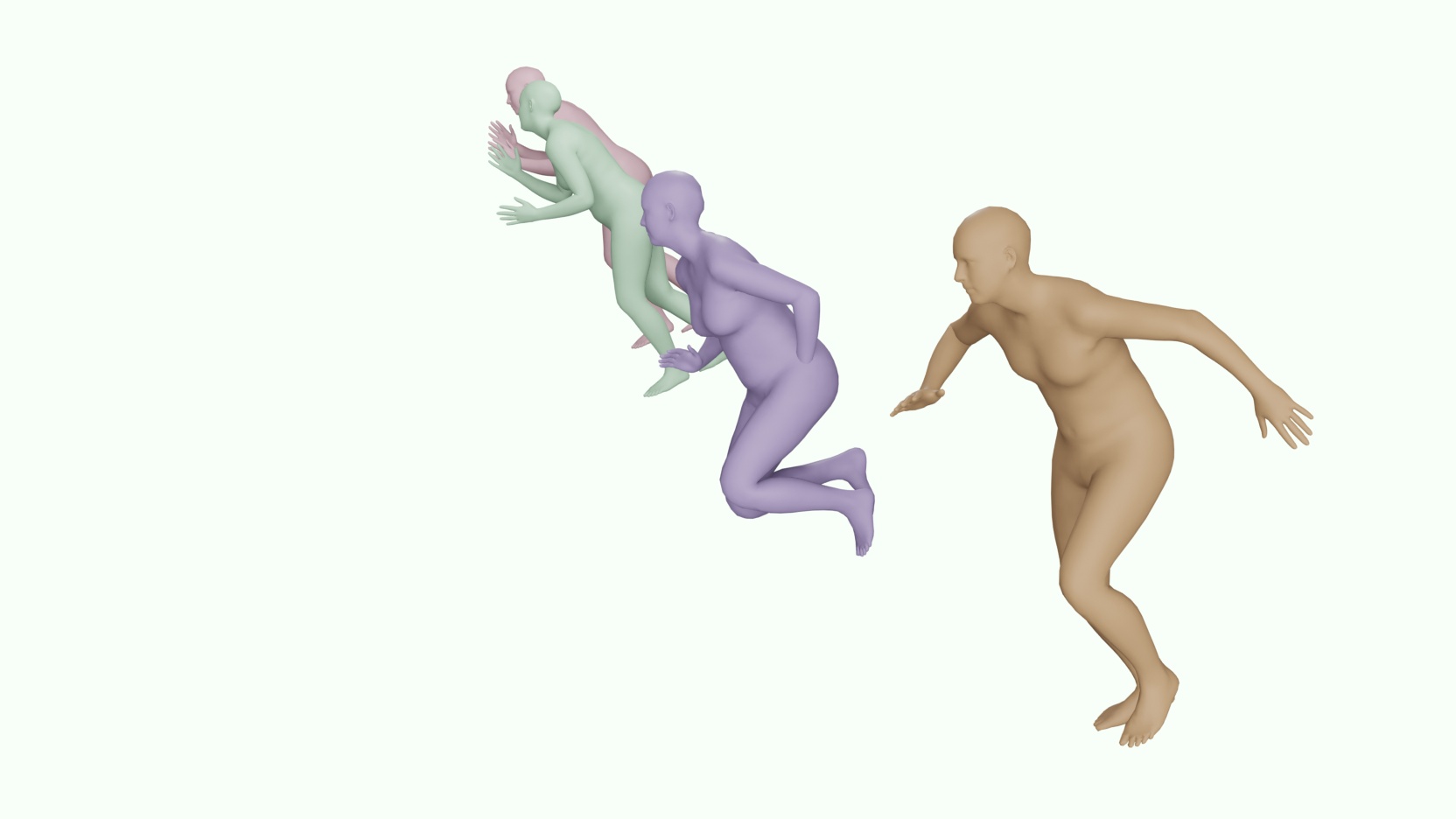}
    \includegraphics[width=0.12\textwidth]{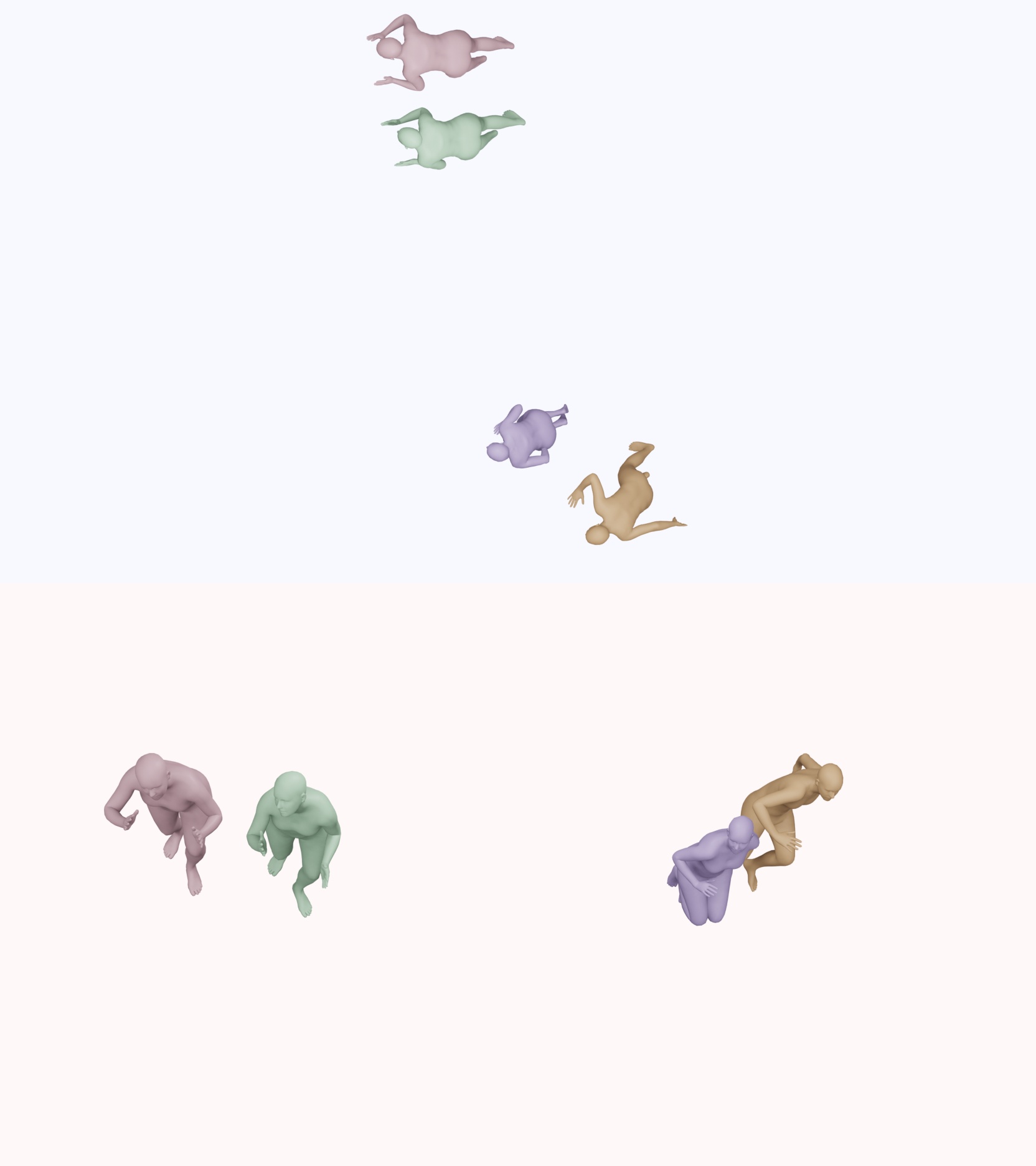}\\
    \includegraphics[width=0.24\textwidth]{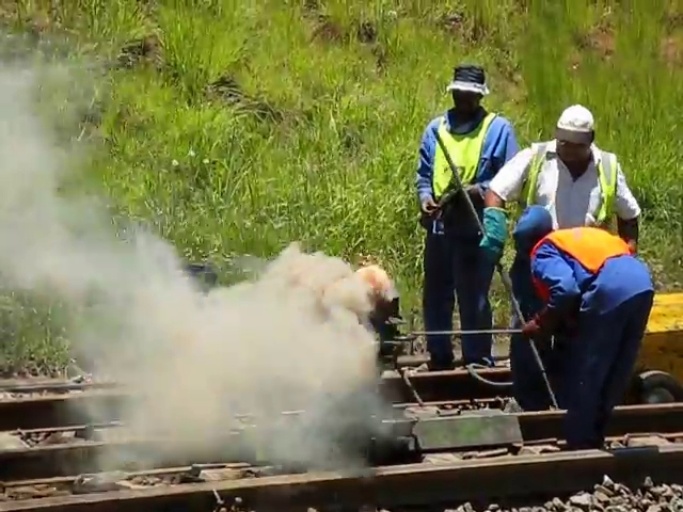}
    \includegraphics[width=0.24\textwidth]{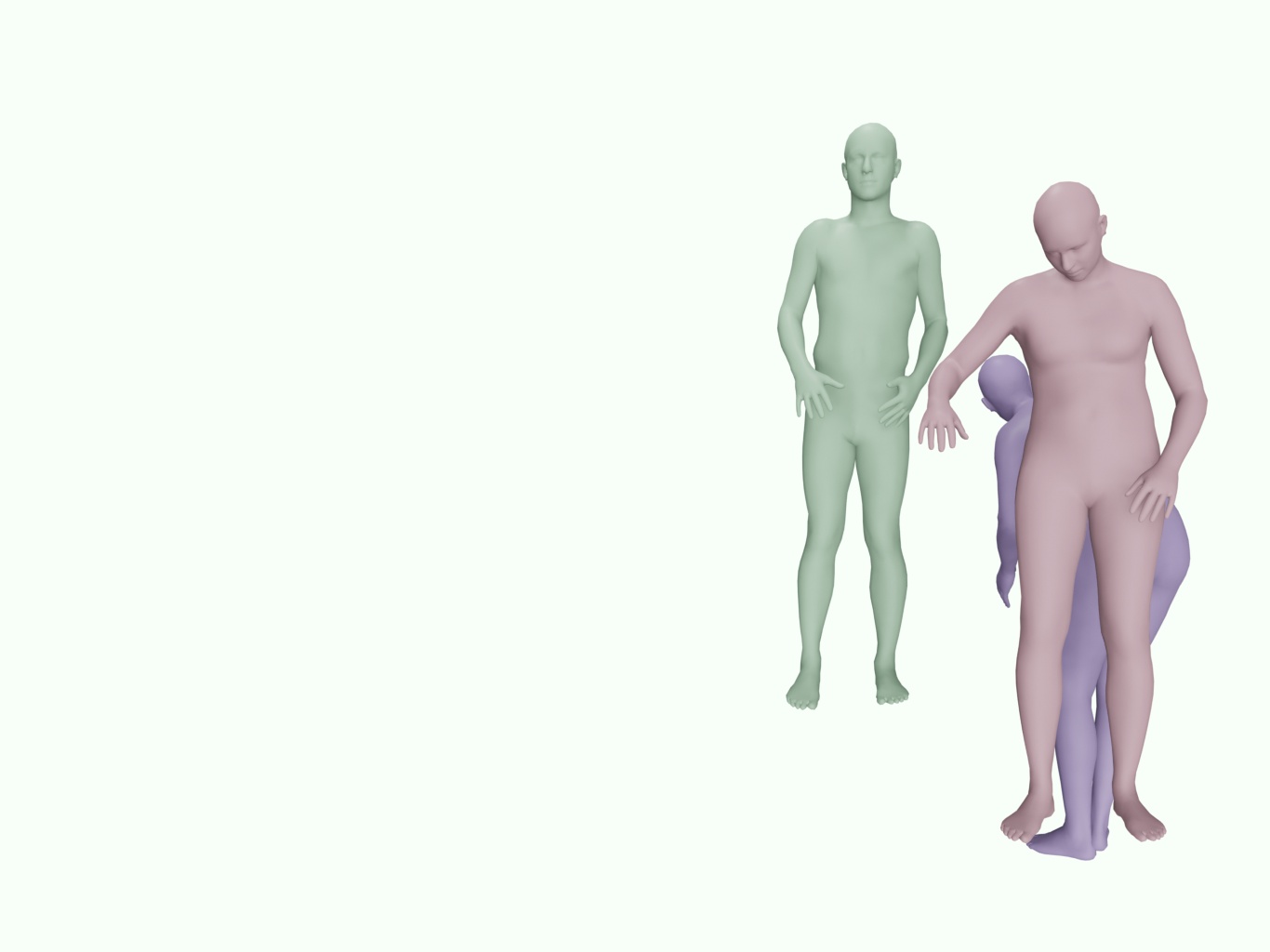}
    \includegraphics[width=0.12\textwidth]{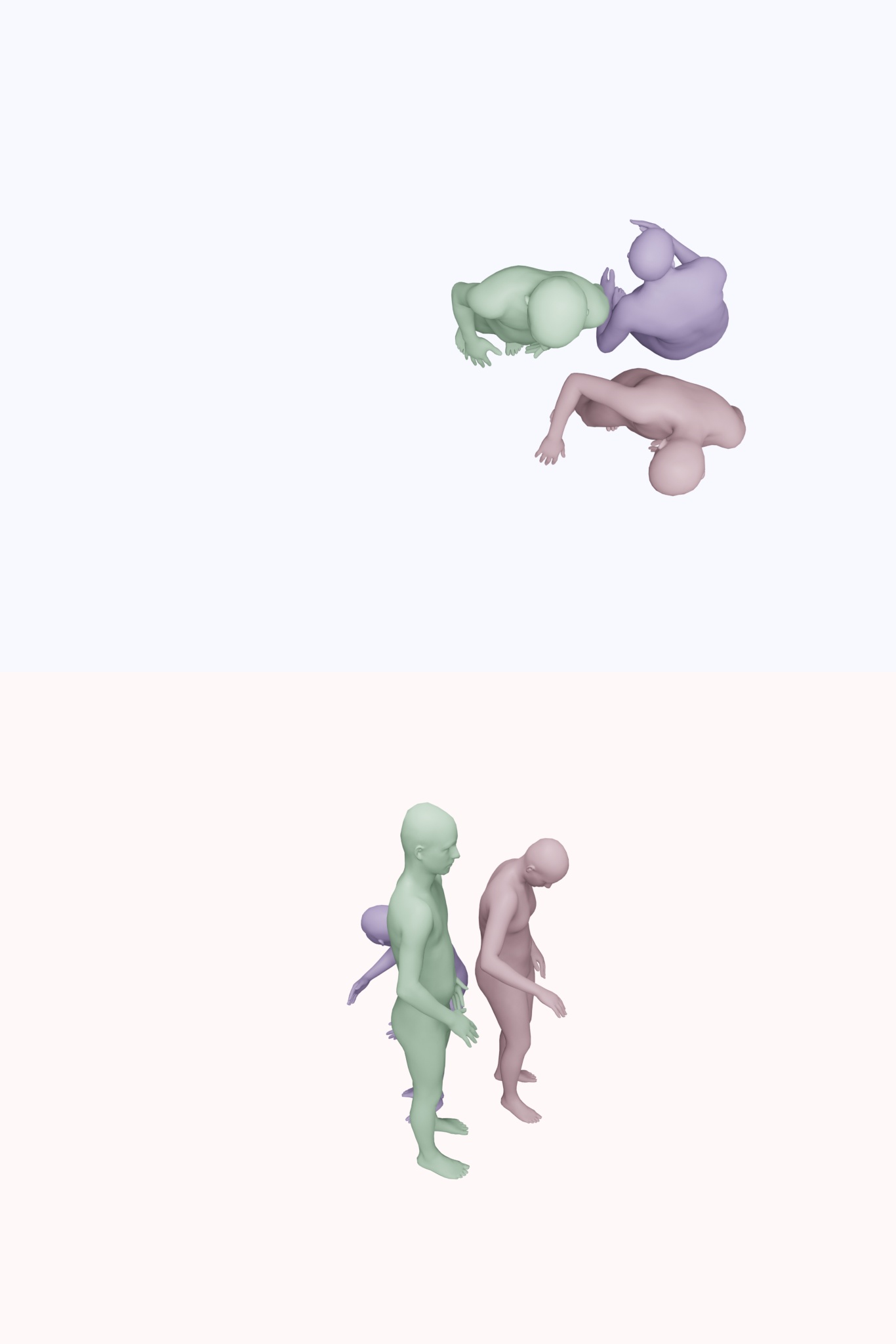}
    \includegraphics[width=0.24\textwidth]{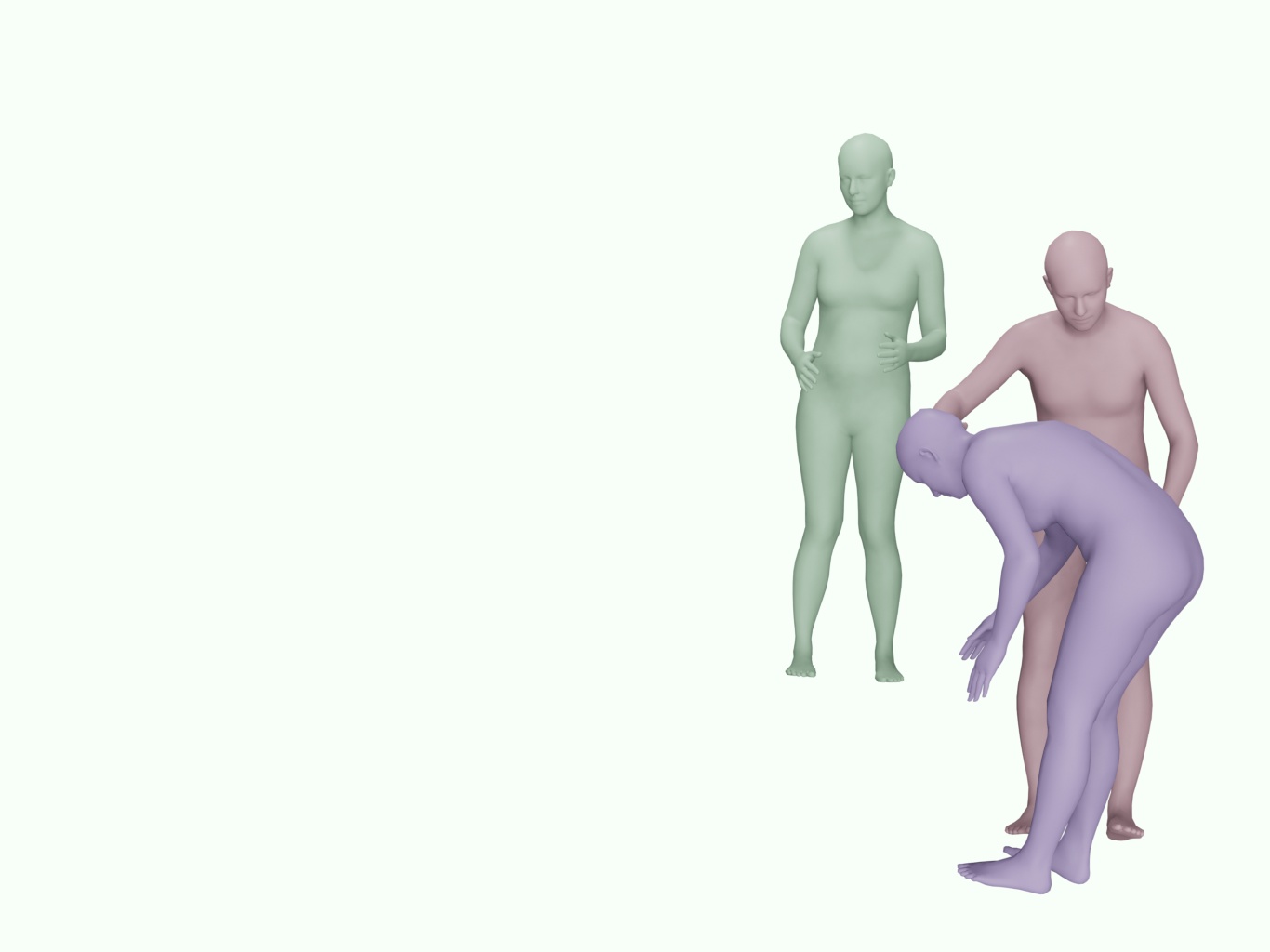}
    \includegraphics[width=0.12\textwidth]{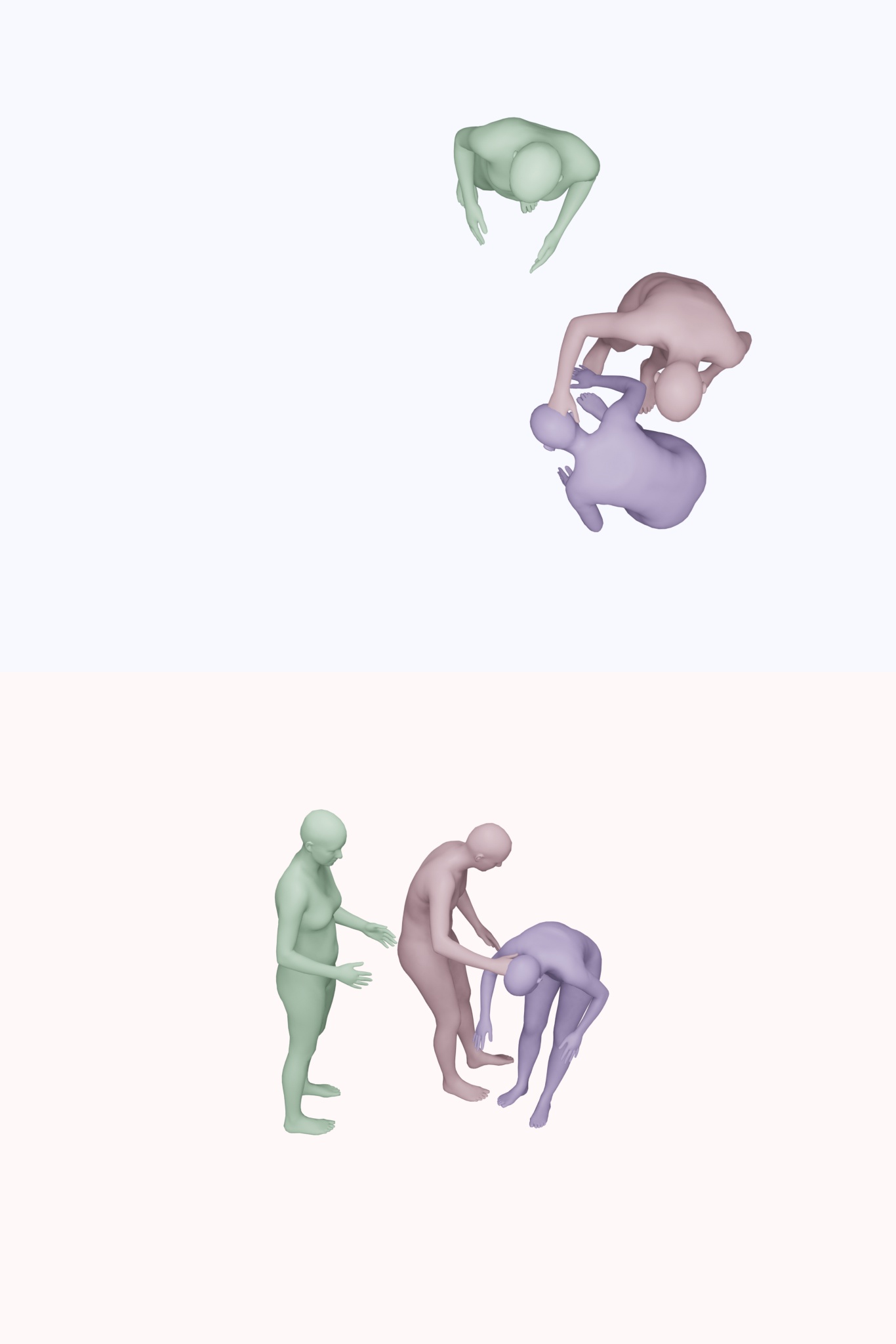}\\
  \vspace{-1mm}
   \text{\scriptsize Input image } \hspace{40mm} \text{\scriptsize Baseline} \hspace{56mm} \text{\scriptsize Ours}  \hspace{12mm} \text{\color{white} \scriptsize}
	\vspace{-3mm}   
	\caption{{\bf Qualitative effect of proposed losses.}
Results of our baseline model (center) and our full model trained with our proposed losses (right). As expected, we improve over our baseline in terms of coherency in the results (i.e., fewer interpenetrations, more consistent depth ordering for the reconstructed meshes).
}
\label{fig:ablative}
\vspace{-2mm}
\end{figure*}

\begin{figure*}[!t]
	\centering
    \includegraphics[width=0.24\textwidth]{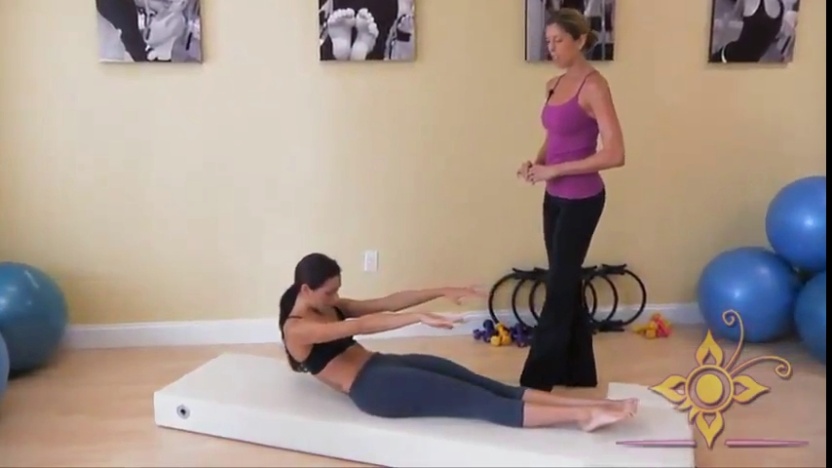}
    \includegraphics[width=0.24\textwidth]{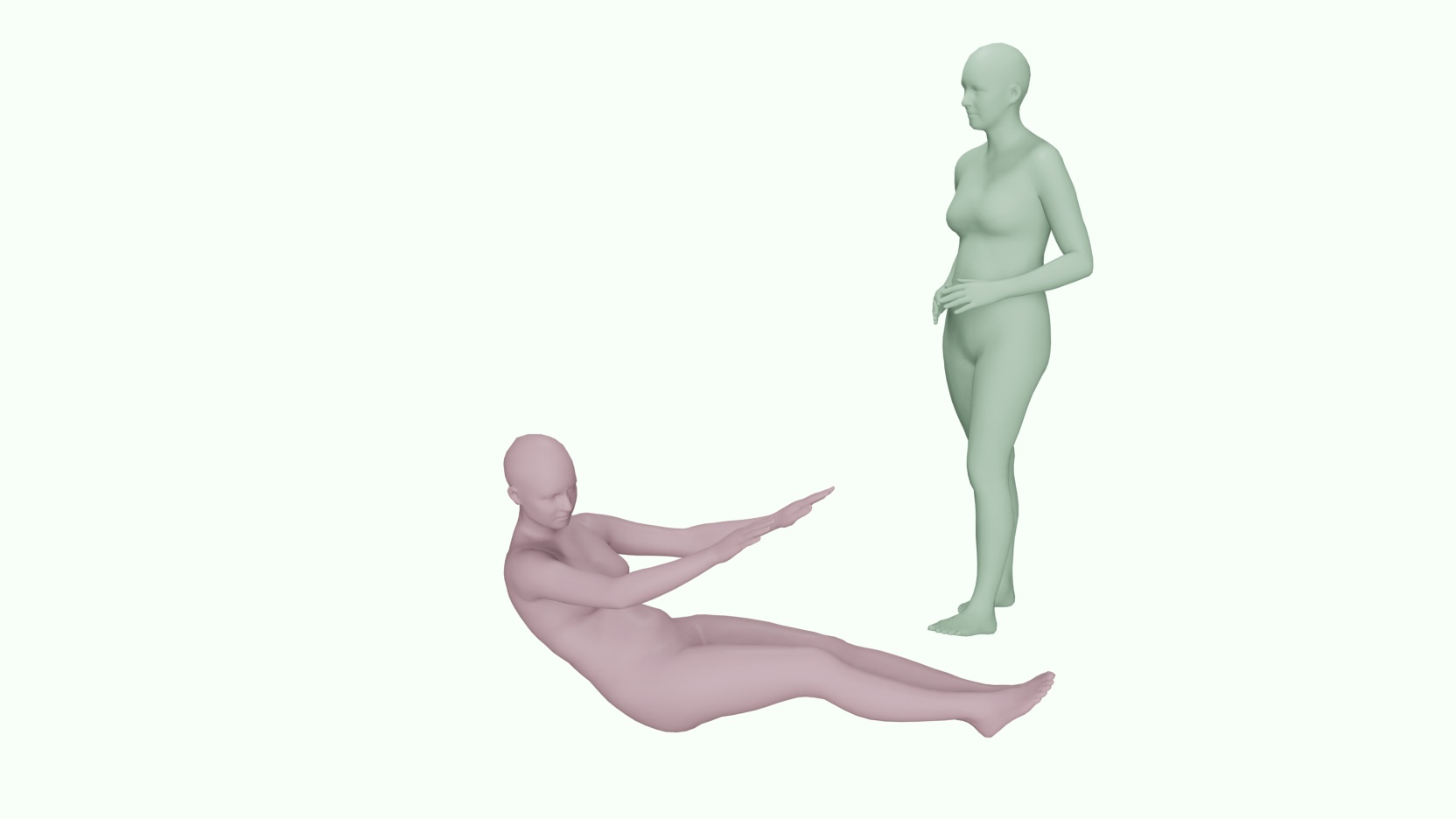}
    \includegraphics[width=0.24\textwidth]{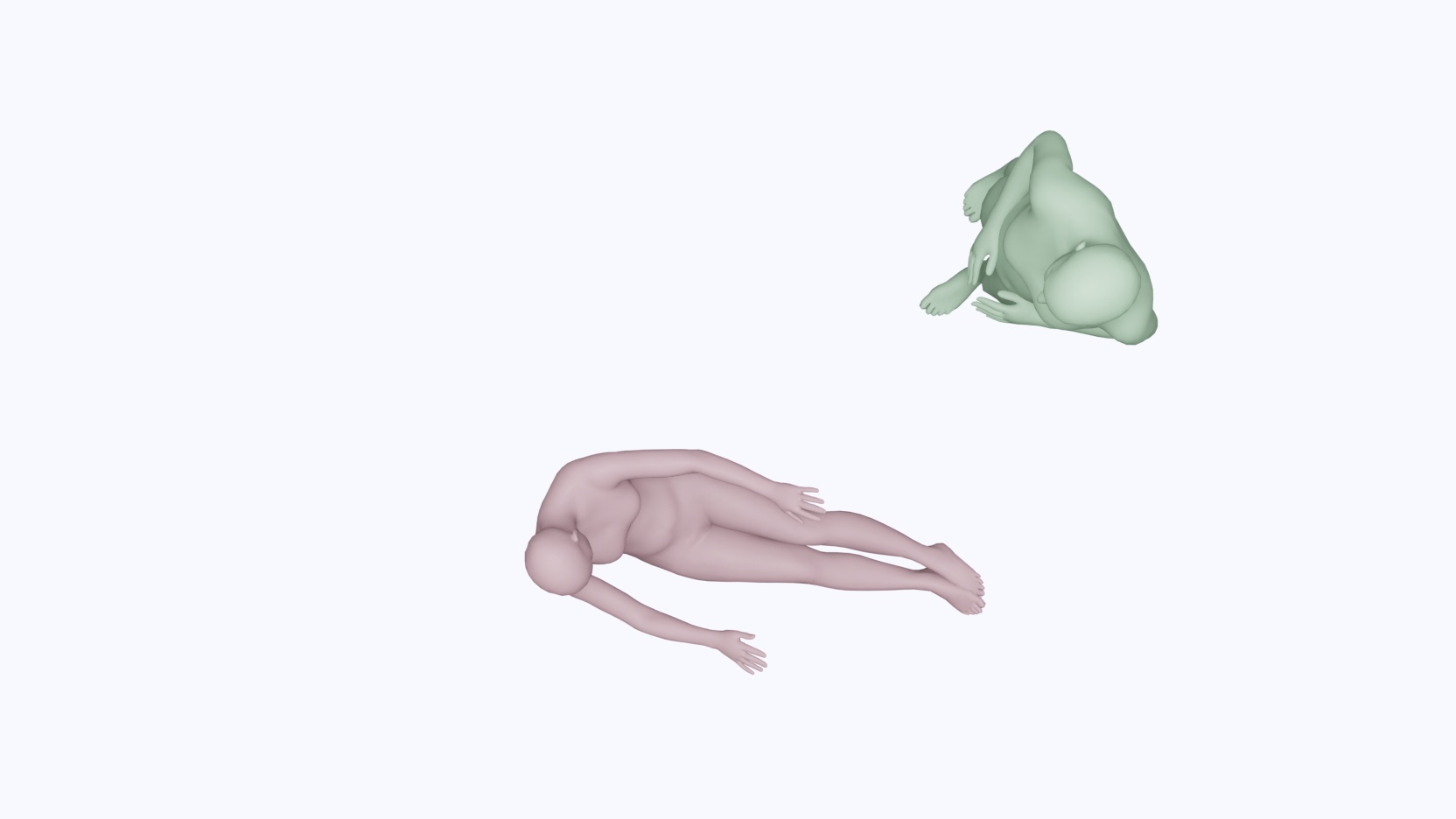}
    \includegraphics[width=0.24\textwidth]{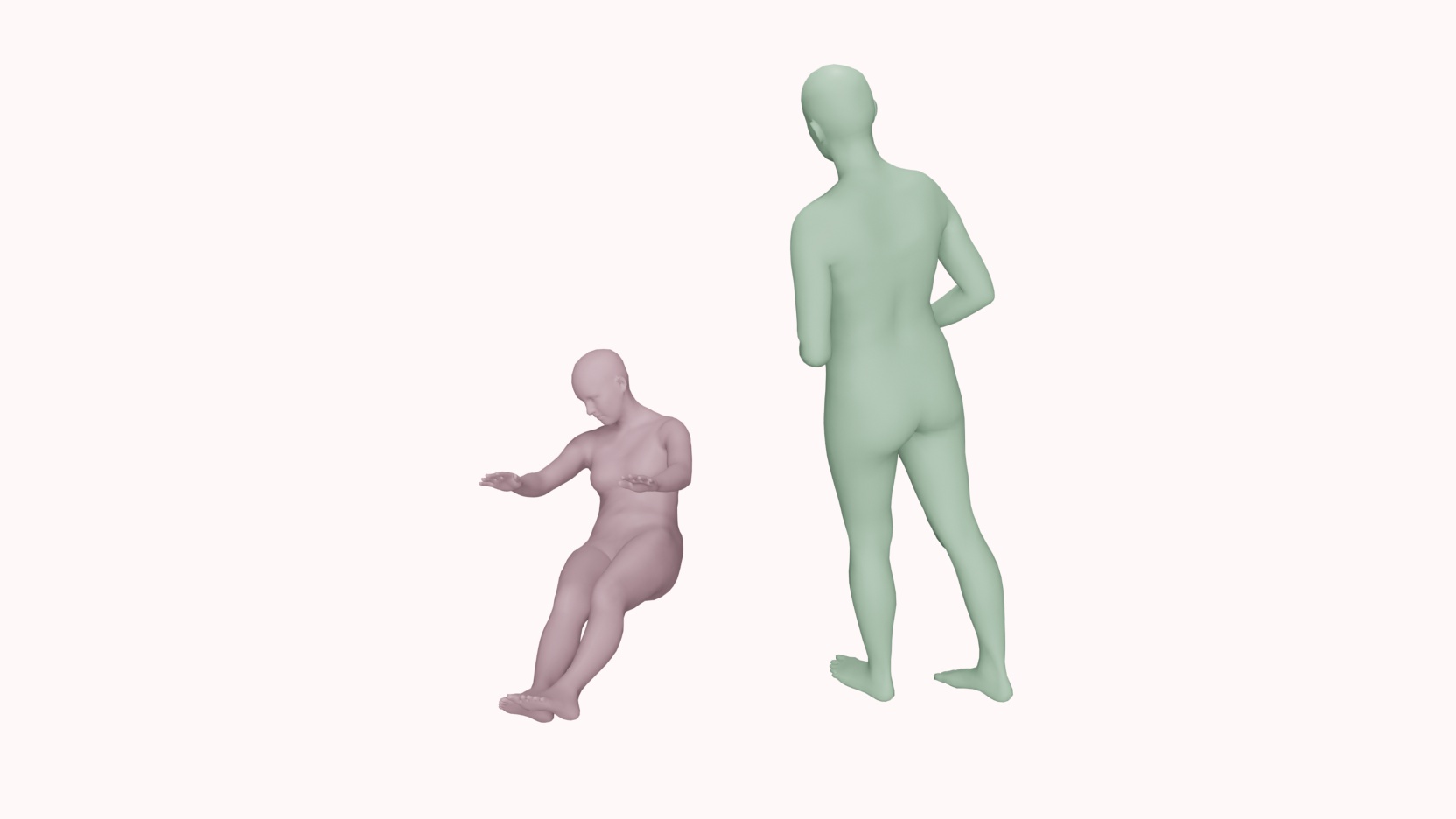}\\
    \includegraphics[width=0.24\textwidth]{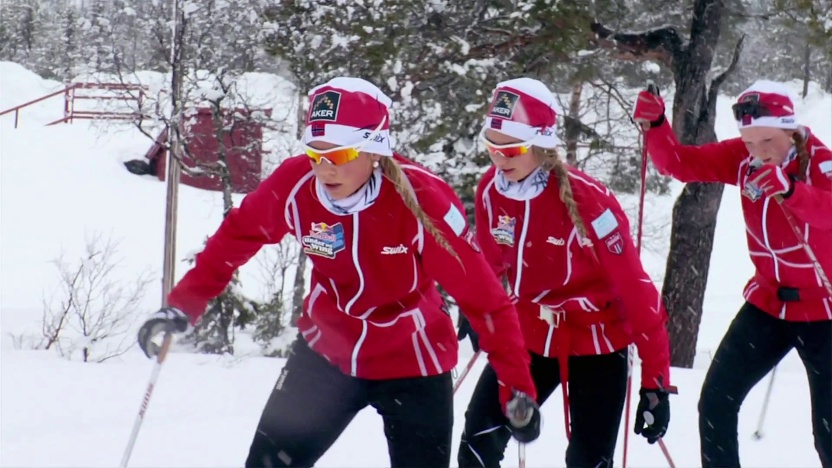}
    \includegraphics[width=0.24\textwidth]{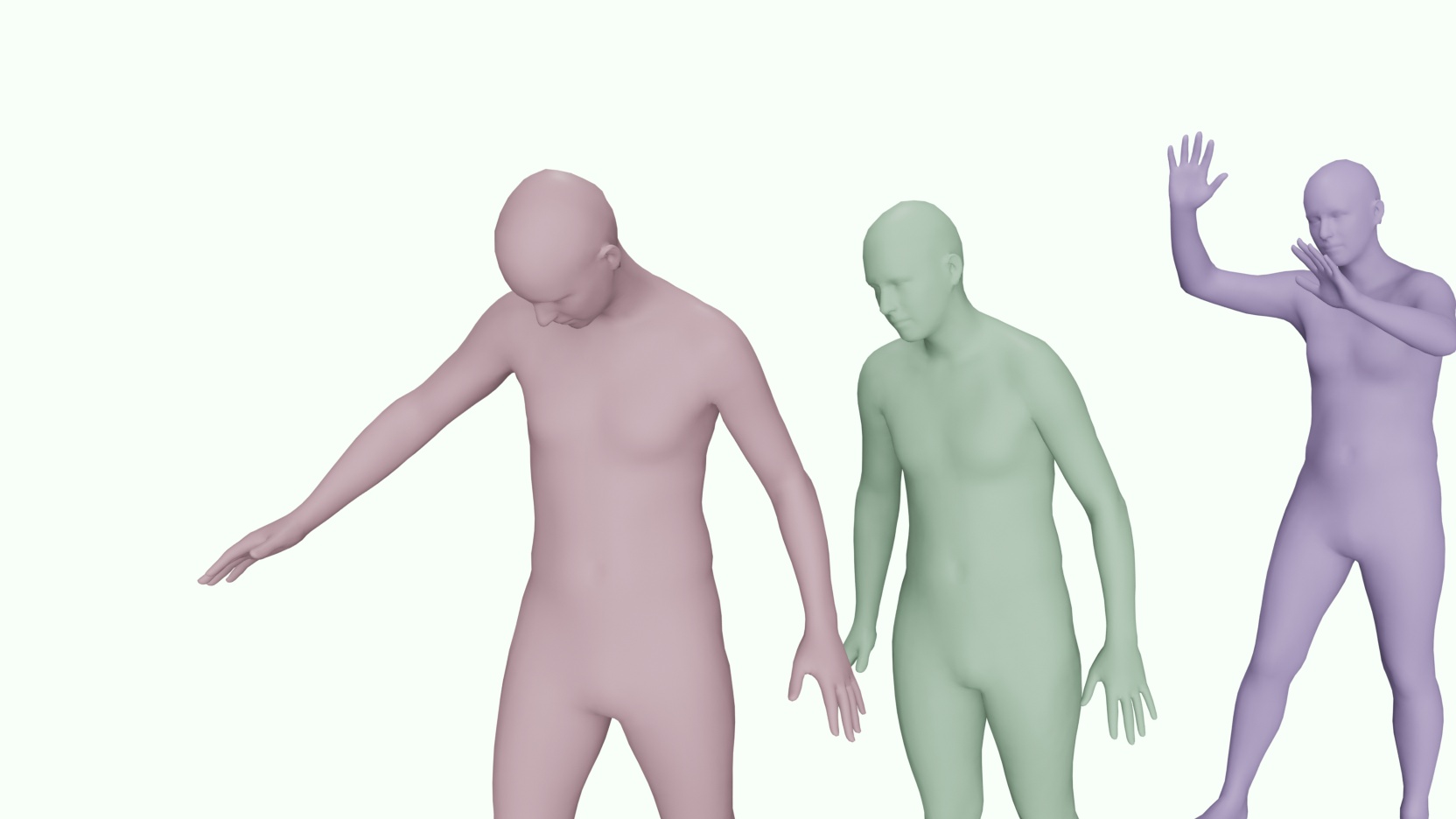}
    \includegraphics[width=0.24\textwidth]{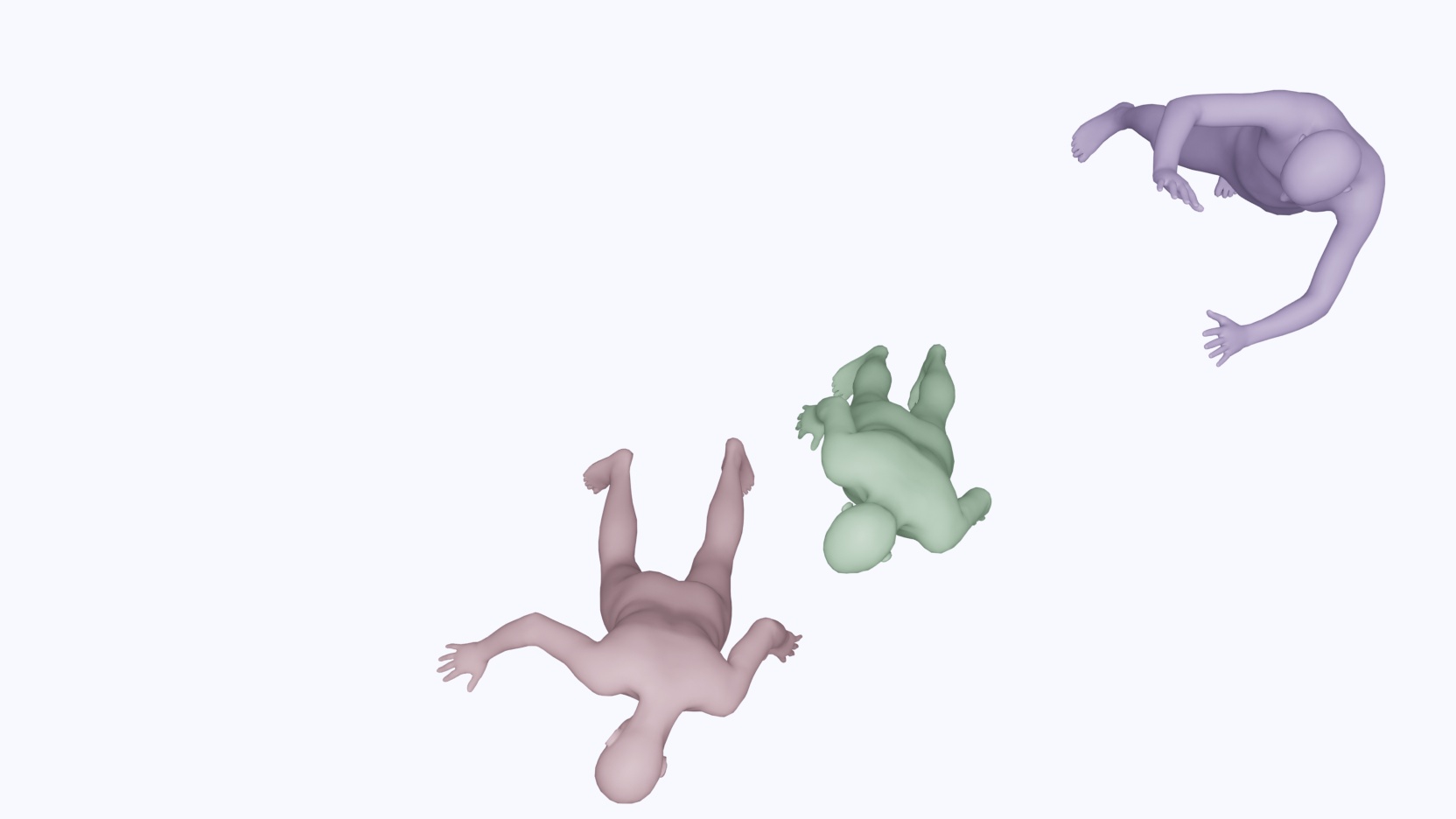}
    \includegraphics[width=0.24\textwidth]{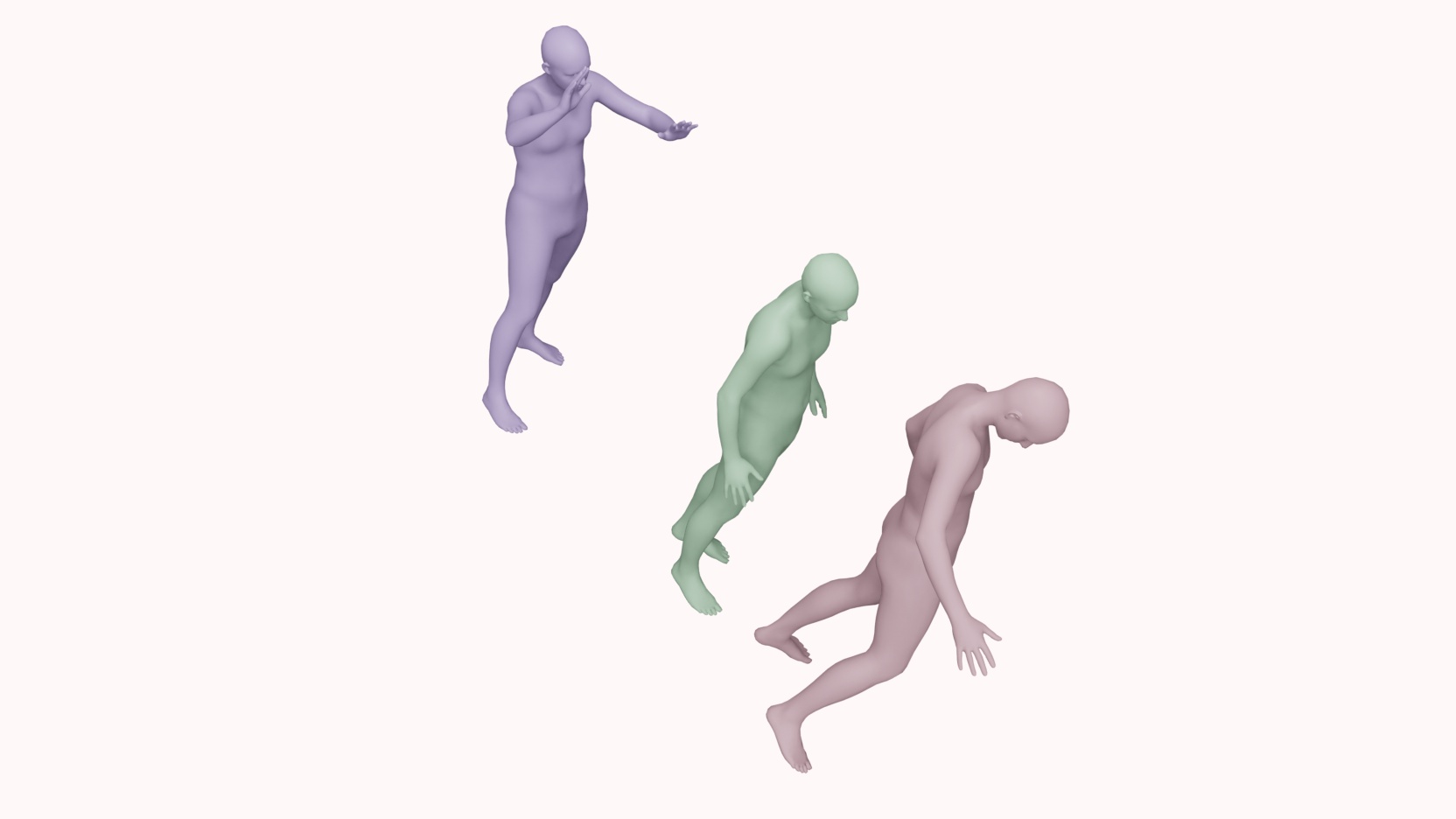}\\
    \includegraphics[width=0.24\textwidth]{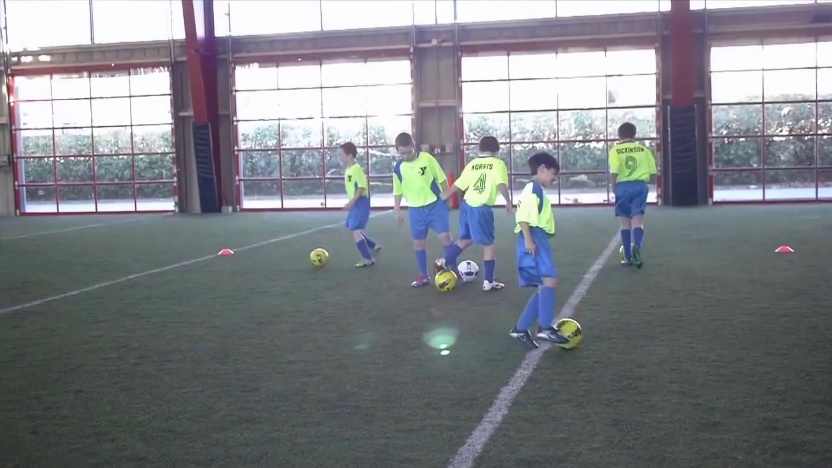}
    \includegraphics[width=0.24\textwidth]{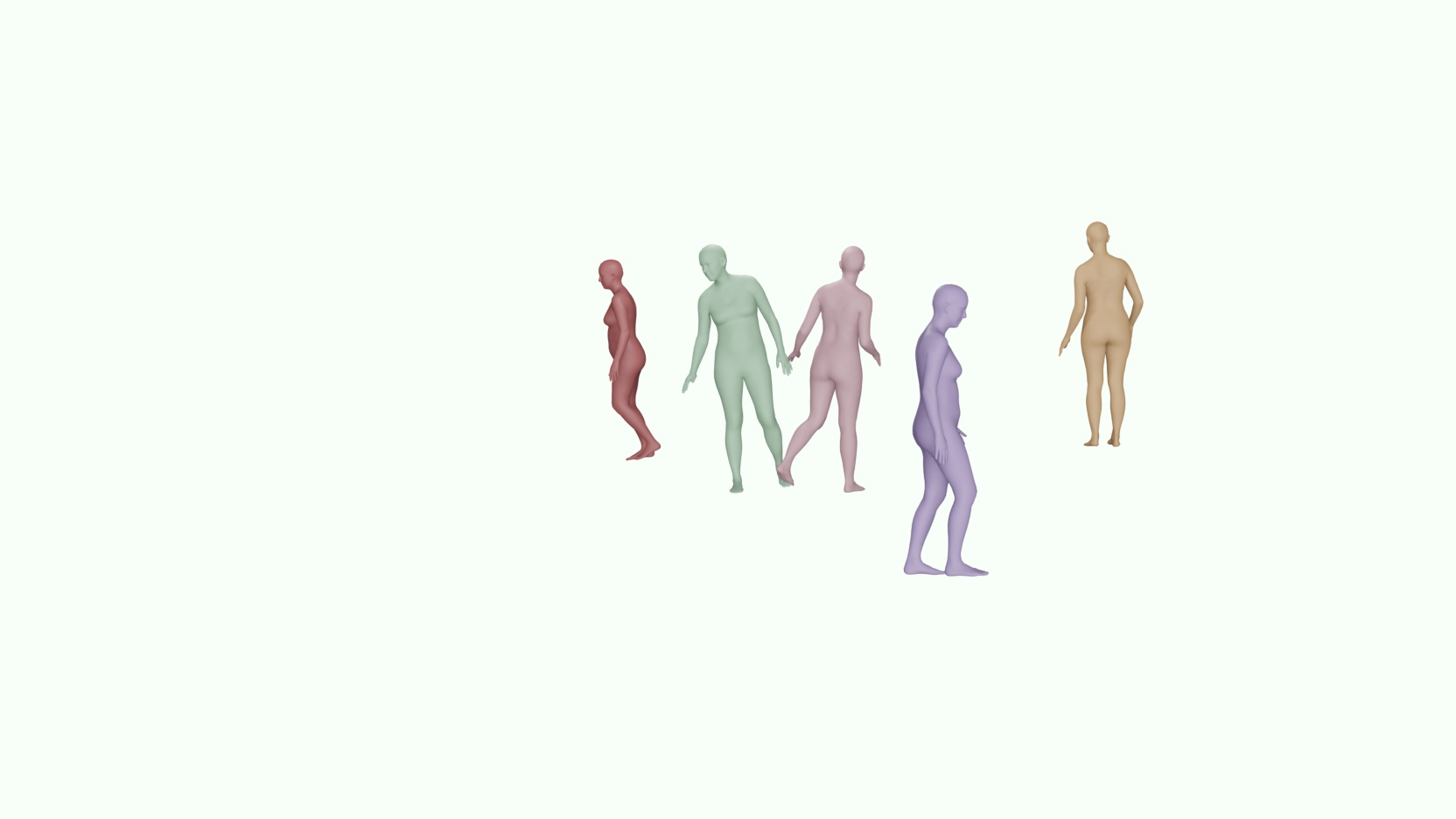}
    \includegraphics[width=0.24\textwidth]{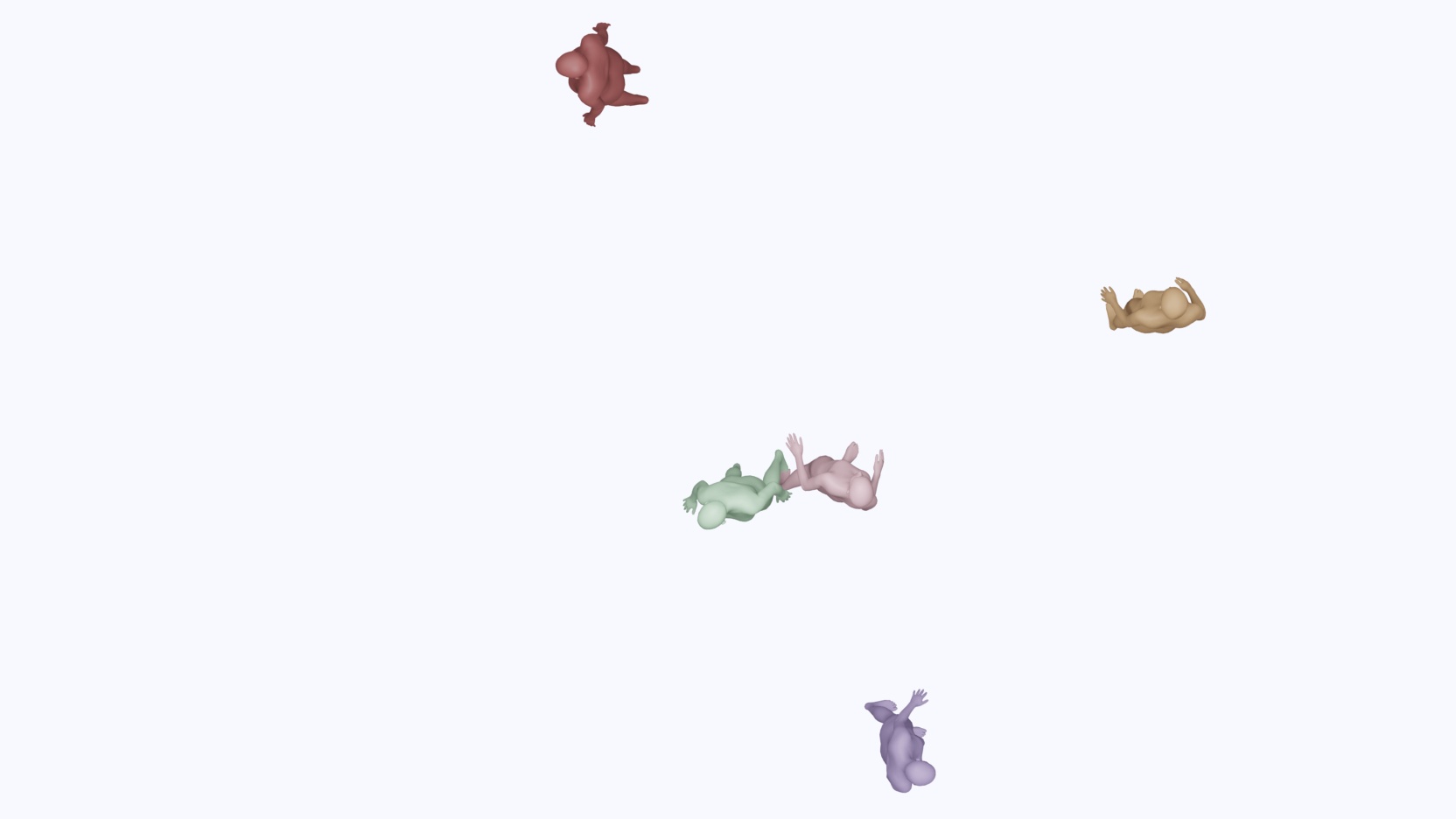}
    \includegraphics[width=0.24\textwidth]{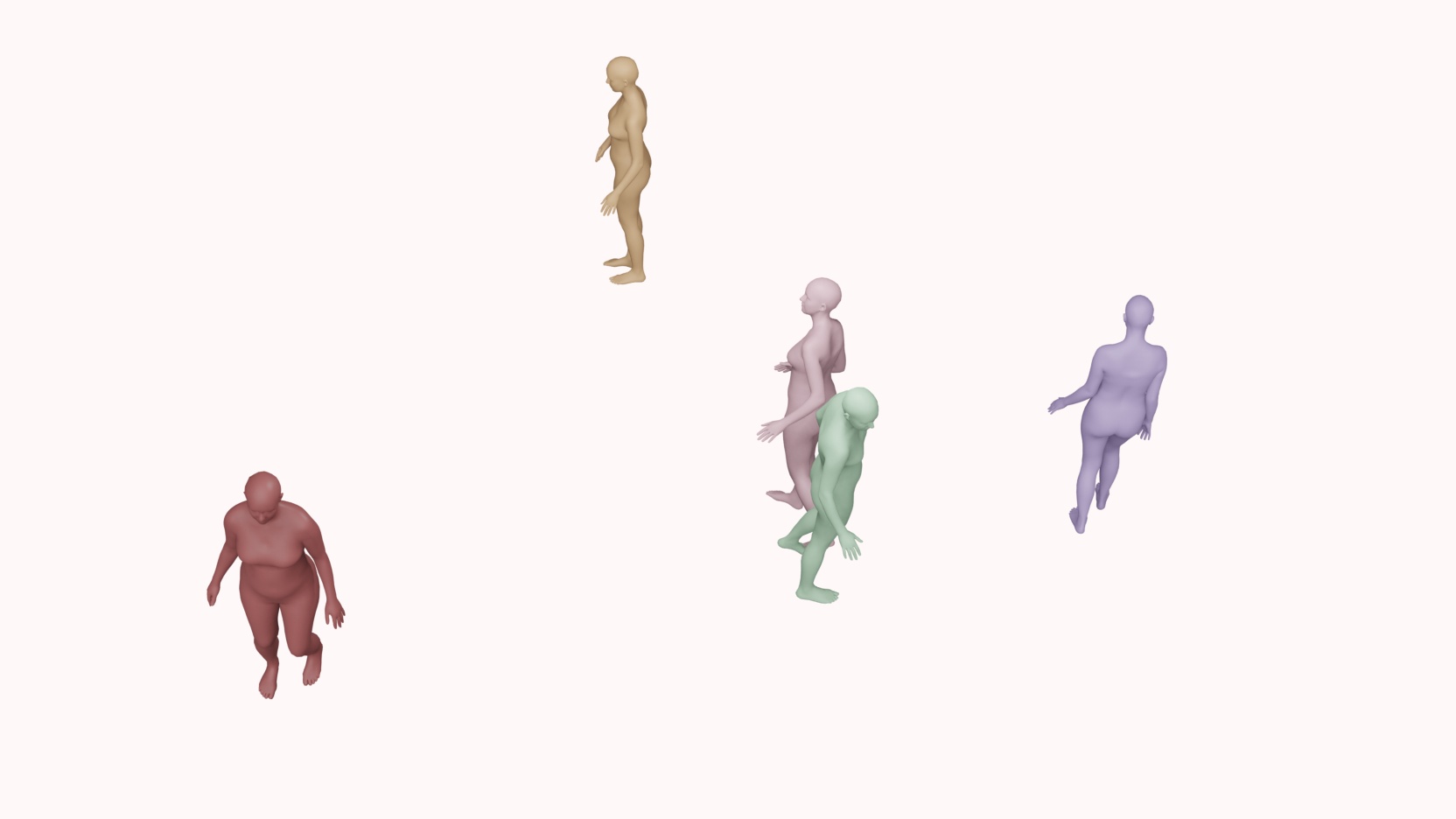}\\
    \includegraphics[width=0.24\textwidth]{figures/final/single/COCO_val2014_000000025393.jpg/img_cropped.jpg}
    \includegraphics[width=0.24\textwidth]{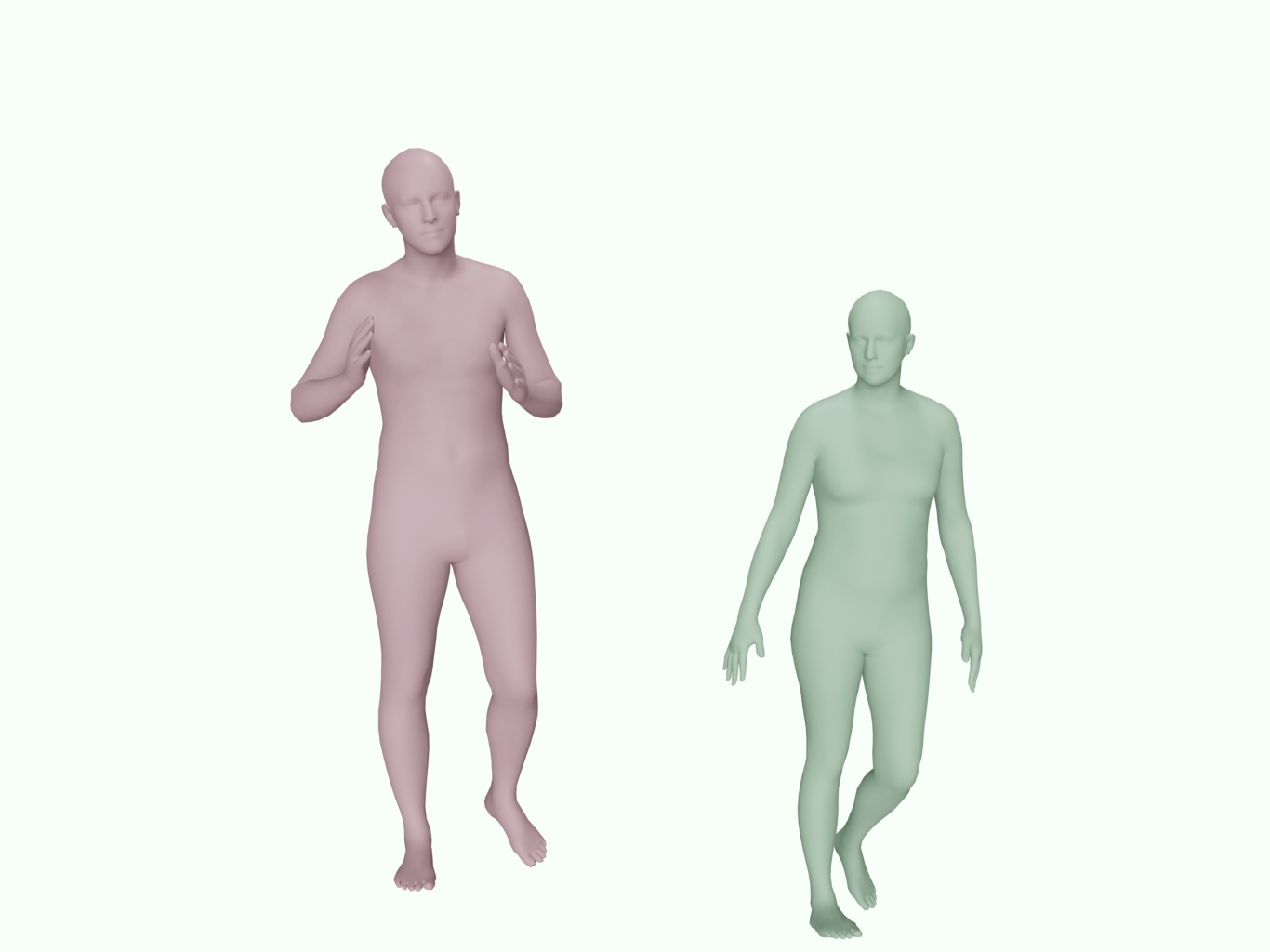}
    \includegraphics[width=0.24\textwidth]{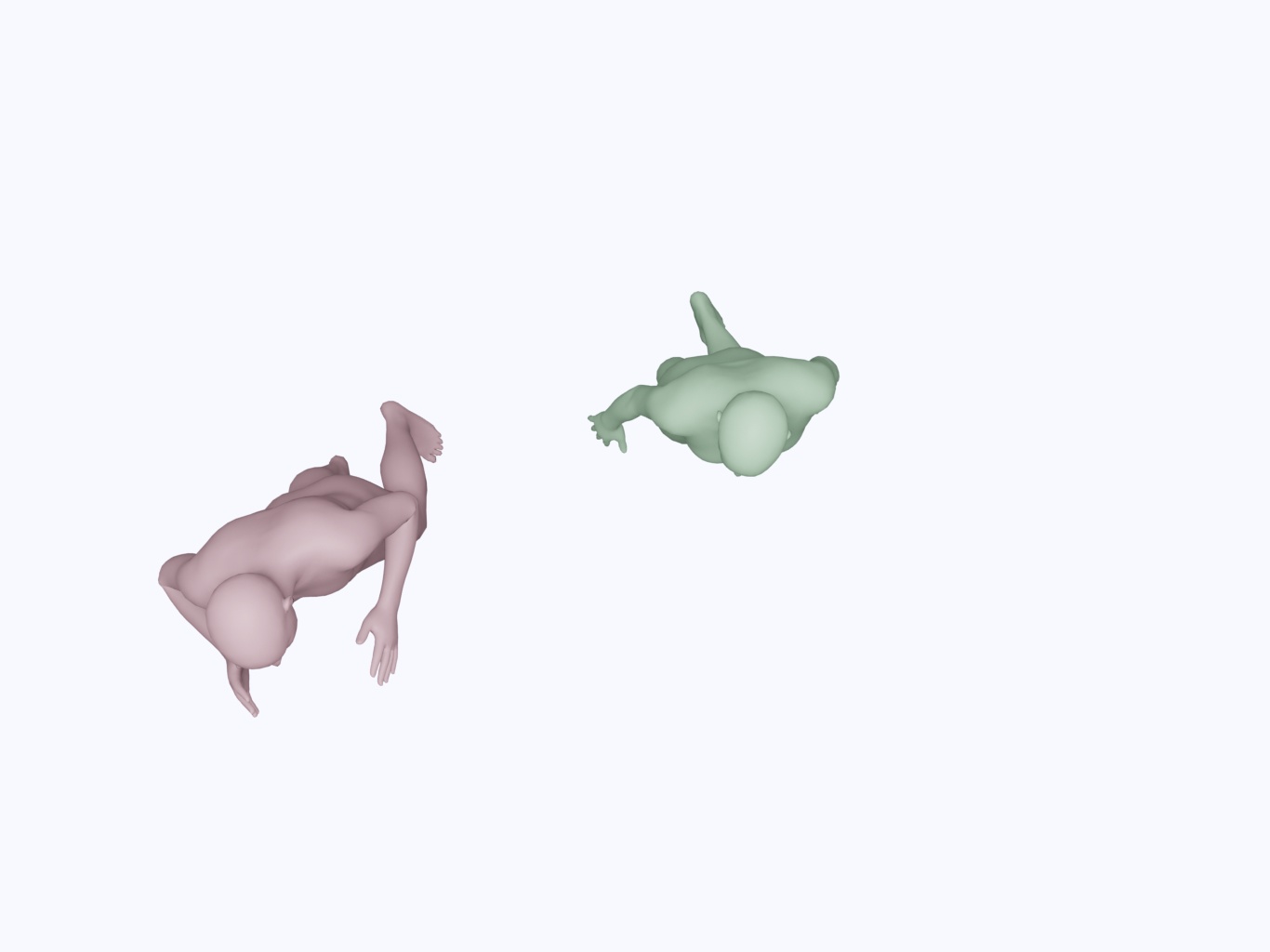}
    \includegraphics[width=0.24\textwidth]{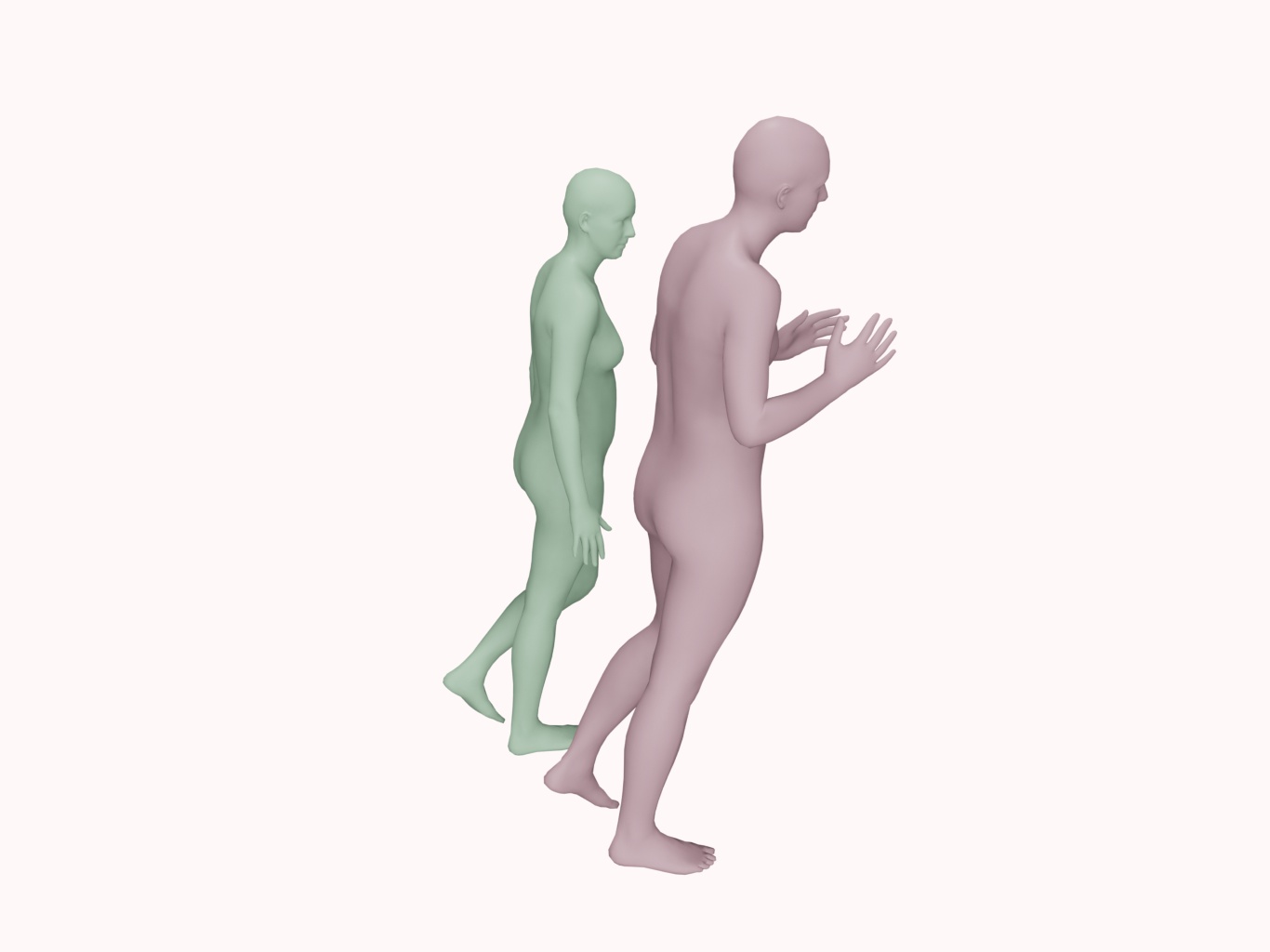}\\ 
	\vspace{-2mm}
	\caption{{\bf Qualitative evaluation.}
We visualize the reconstructions of our approach from different viewpoints; front (green background), top (blue background) and side (red background). More qualitative results can be found in the Sup.Mat.
}
\label{fig:qualitative}
\vspace{-2mm}
\end{figure*}

\subsection{Qualitative evaluation} \label{sec:qualitative}
In this Subsection, we present more qualitative results of our approach. In Figure~\ref{fig:ablative} we compare our baseline with our full model trained with the proposed losses. As expected, our full model generates more coherent reconstructions, improving over the baseline as far as interpenetration and depth ordering mistakes are concerned. Errors can happen when there is significant scale difference among the people and there is no overlap on the image plane (last row of Figure~\ref{fig:qualitative}). More results can be found in the Sup.Mat.

\section{Summary}
In this work, we present an end-to-end approach for multi-person 3D pose and shape estimation from a single image. Using the R-CNN framework, we design a top-down approach that regresses the SMPL model parameters for each detected person in the image. Our main contribution lies on assessing the problem from a more holistic view and aiming on estimating a coherent reconstruction of the scene instead of focusing only on independent pose estimation for each person. To this end, we incorporate two novel losses in our framework that train the network such that a) it avoids generating overlapping humans and b) it is encouraged to position the people in a consistent depth ordering. We evaluate our approach in various benchmarks, demonstrating very competitive performance in the traditional 3D pose metrics, while also performing significantly better both qualitatively and quantitatively in terms of coherency of the reconstructed scene. In future work, we aim to more explicitly model interactions between people (besides the overlap avoidance), so that we can achieve a more accurate and detailed reconstruction of the scene at a finer level as well. In a similar vein, we can incorporate further information towards a holistic reconstruction of scenes. This can include constraints from the ground plane~\cite{zanfir2018monocular}, background~\cite{hassan2019resolving}, or the objects that humans interact with~\cite{hasson2019learning,tekin2019h+}.

\footnotesize
\noindent
{\bf Acknowledgements:} 
NK, GP and KD gratefully appreciate support through the following grants: 
NSF-IIP-1439681 (I/UCRC), NSF-IIS-1703319, NSF MRI 1626008, ARL RCTA W911NF-10-2-0016, ONR N00014-17-1-2093, ARL DCIST CRA W911NF-17-2-0181, the DARPA-SRC C-BRIC, by Honda Research Institute and a Google Daydream Research Award. XZ and WJ would like to acknowledge support from NSFC (No. 61806176) and Fundamental Research Funds for the Central Universities (2019QNA5022). 

{\small
\balance
\bibliographystyle{ieee_fullname}
\bibliography{egbib}
}

\end{document}